\documentclass[times,twocolumn,final]{elsarticle}

\usepackage{amssymb}
\usepackage{amsmath}
\usepackage{cag}
\usepackage{booktabs}
\usepackage{comment}
\usepackage{framed,multirow}
\usepackage{color}

\usepackage{latexsym}

\usepackage{url}
\usepackage{xcolor}
\definecolor{newcolor}{rgb}{.8,.349,.1}

\usepackage{hyperref}

\journal{Computers \& Graphics}

\begin{document}

\verso{Preprint Submitted for review}

\begin{frontmatter}

\title{Learning Modified Indicator Functions for Surface Reconstruction}%

\author[1] {Dong Xiao}
\ead{xiaod18@mails.tsinghua.edu.cn}

\author[2]{Siyou Lin}
\ead{linsy21@mails.tsinghua.edu.cn}

\author[2,4]{Zuoqiang Shi\corref{cor}}
\ead{zqshi@tsinghua.edu.cn}

\author[1,3] {Bin Wang\corref{cor}}
\ead{wangbins@tsinghua.edu.cn}
\cortext[cor]{Corresponding authors}

\address[1]{School of Software, Tsinghua University, Beijing, China}
\address[2]{Department of Mathematical Sciences, Tsinghua University, Beijing, China}
\address[3]{Beijing National Research Center for Information Science and Technology, Beijing, China}
\address[4]{Yanqi Lake Beijing Institute of Mathematical Sciences and Applications, Beijing, China}


\begin{abstract}

Surface reconstruction is a fundamental problem in 3D graphics. In this paper, we propose a learning-based approach for implicit surface reconstruction from raw point clouds without normals. Our method is inspired by Gauss Lemma in potential energy theory, which gives an explicit integral formula for the indicator functions. We design a novel deep neural network to perform surface integral and learn the modified indicator functions from un-oriented and noisy point clouds. We concatenate features with different scales for accurate point-wise contributions to the integral. Moreover, we propose a novel Surface Element Feature Extractor to learn local shape properties. Experiments show that our method generates smooth surfaces with high normal consistency from point clouds with different noise scales and achieves state-of-the-art reconstruction performance compared with current data-driven and non-data-driven approaches. The source code is available at \url{https://github.com/Submanifold/LMIRecon}.

\end{abstract}


\end{frontmatter}

\section{Introduction}

Surface reconstruction has been extensively studied for more than three decades due to its rich applications~\cite{ReconSurvey2017}. Many elegant classical (non-data-driven) implicit approaches can generate high-fidelity surfaces when the input contains orientated normals with global consistency~\cite{2005Recon, 2006Poisson, 2011SSD, 2013Screened, 2020EnvelopePoisson, 2019GaussRecon}. However, it is not a trivial task to obtain accurately oriented normals from raw point clouds. What is more, non-data-driven methods usually do not have a mechanism to distinguish noise. Thus, their performance degenerates when the scanning noise is large or the per-point normal estimation is inaccurate.

\begin{figure*}[htb]
  \centering
  \includegraphics[width=1.0\linewidth]{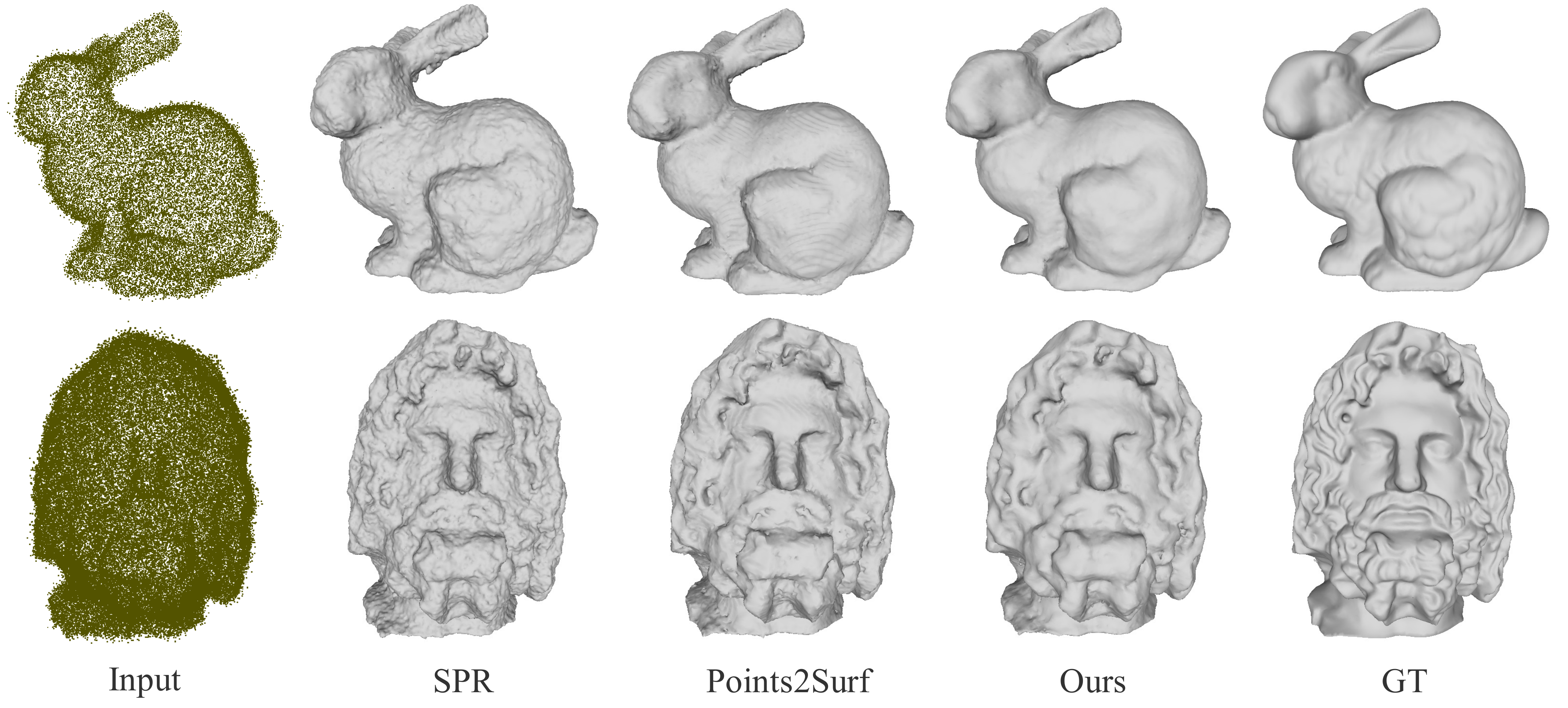}
  \caption{
           Comparison of our method with Screened Poisson Surface Reconstruction (SPR)~\cite{2013Screened} and Points2Surf~\cite{2020Points2Surf}. They are representative approaches for non-data-driven and data-driven reconstruction, respectively. SPR tends to overfit the noisy points when the input point cloud is unclean. Points2Surf may produce bumpy surfaces with ripples. Our method can generate smooth surfaces with better normal consistency from raw and noisy inputs.}
    \label{fig:Figure1}
\end{figure*}

In recent years, deep learning has gradually extended to increasing 3D geometric problems such as point cloud classification, segmentation, normal estimation, and surface reconstruction~\cite{2017PointNet, 2017PointNet++,2018PointCNN, 2019PointConv, 2018PCPNet, 2018AtlasNet, 2019Occupancy, 2020Points2Surf, 2020ConvOccupancy}. Previous studies indicate that deep neural networks can be good representations of implicit functions, and some data-driven methods for implicit representation have been proposed ~\cite{2019Occupancy, 2019DeepSDF, 2019Scan2Mesh}. However, many of them are data-specific, specializing in particular categories, and have poor generalizability. Reconstructing high-quality surfaces from arbitrary un-oriented point clouds with complex geometric topology remains a challenging problem.

Points2Surf~\cite{2020Points2Surf} is an outstanding and representative work to directly estimate the implicit functions for arbitrary point clouds without normals. It learns the absolute distance through a detailed local patch and learns the sign from a global sampling. Points2Surf has good generalizability and adapts to different noise scales. However, it may produce bumpy surfaces with ripples and cause a decrease in normal consistency when we observe the reconstruction details.

\begin{figure}[htb]
  \centering
  \includegraphics[width=1.0\linewidth]{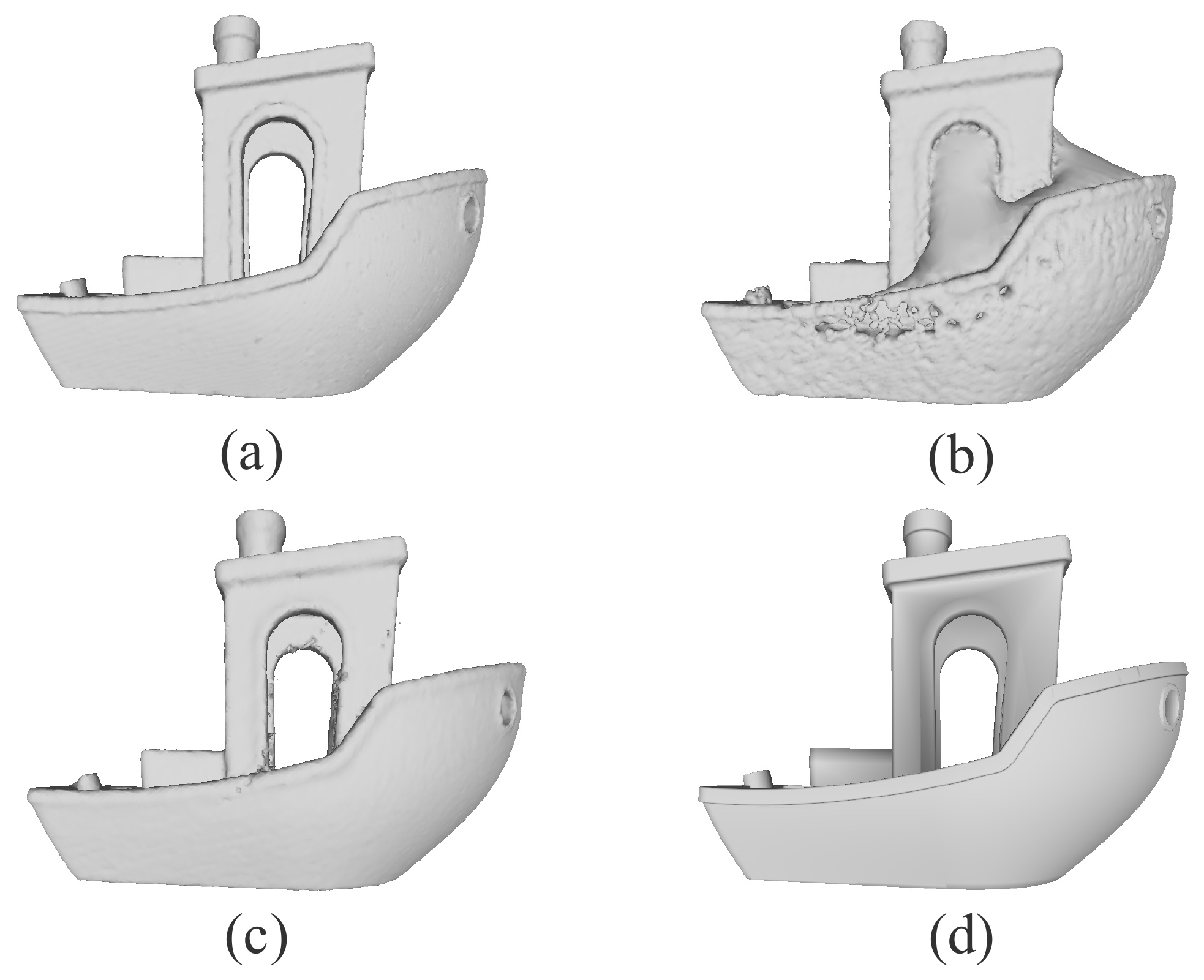}
  \caption{
           Comparison between our method and GR~\cite{2019GaussRecon}. (a) GR with fully accurate normals sampled directly from the ground truth surface. (b) GR with inaccurate normals and an inconsistent orientation estimated by PCPNet~\cite{2018PCPNet}. (c) Our reconstruction from the raw point cloud without normals. (d) Ground truth.}
    \label{fig:Figure2}
\end{figure}

Motivated by Gauss lemma in potential energy theory~\cite{1995PDE, 2009Potential}, we propose a method to directly learn modified indicator functions from raw point clouds. Gauss lemma indicates that the indicator value can be expressed as an explicit surface integral when given a proper kernel function. GR~\cite{2019GaussRecon} is a classical reconstruction method based on Gauss lemma. It can generate high-quality surfaces with clean input and avoid overfitting to noisy sample points to a certain extent. However, this method requires globally oriented normals and relies heavily on the normal accuracy, especially orientation consistency. As shown in Fig.~\ref{fig:Figure2}, GR can generate a high-quality surface when the sampling is clean and the input normals are accurate. However, its performance degrades greatly when the normal estimation is inaccurate. By contrast, our method can generate smooth and consistent reconstructions under raw point clouds. Fig.~\ref{fig:Figure1} shows the comparison of our method with representative data-driven and non-data-driven approaches. Our method exhibits better normal consistency under raw and noisy inputs.

If we use the traditional way to process the surface integral in Gauss lemma as GR, we will encounter the same severe singularity problems near the surface. That is, the integrand tends to infinity. A complicated refinement process is always required for approximation. Even so, the noise and the inaccurate normals still affect the results. Thus, we leverage the powerful learning ability of neural networks. We propose a novel Surface Element Feature Extractor (SEF Extractor), which aggregates local shape properties for accurate calculation in the singular area. Performing global surface integral by the neural network is also a novel problem. In this work, we process it by aggregating the point-wise contributions in the network. To obtain accurate point-wise information, we concatenate features with different scales from local to global latent. The indicator prediction is obtained by the sum of the point-wise values. Through this approach, we have access to an effective mechanism for supervising the global latent vector and distributing the potential errors to sample points so as to produce robust and smooth reconstructions.

We conduct sufficient experiments on different datasets with varying noise, different sampling densities and artifacts. The results indicate that our method outperforms the existing data-driven and non-data-driven approaches on reconstruction from un-oriented point clouds, especially with greater advantage in normal consistency.

\section{Related Works} \label{related_work}

Surface reconstruction from point clouds is an extensively studied topic in computer graphics. Many outstanding algorithms are available. In this section, we briefly review several existing data-driven and non-data-driven implicit-based approaches that are most relevant to us.

\subsection{Implicit representation}
An implicit function $f : \mathbb{R}^3 \rightarrow \mathbb{R}$ is defined in three-dimensional Euclidean space and the surface can be presented as a level set of the implicit function. This function is always encoded on a volumetric grid or an octree, and can also be represented by a neural network~\cite{2019Occupancy, 2019DeepSDF}. The commonly used implicit functions include indicator functions~\cite{2006Poisson, 2013Screened, 2019GaussRecon} and signed distance functions~\cite{1992UnorganizedRecon, 2011SSD, 2020Points2Surf}. The traditional indicator function has value zero outside the surface and value one inside. Thus, it is discontinuous near the surface. To address this problem, Kazhdan et al.~\cite{2006Poisson, 2013Screened} apply a smooth filter to the normal field. Lu et al.~\cite{2019GaussRecon} utilize a modified Gauss formula to construct a relatively smooth function near the surface. Different from these traditional methods, we use a neural network to directly learn the modified indicator functions with good properties.

\subsection{Non-data-driven reconstruction}
Non-data-driven surface reconstruction has a history of more than three decades. It often requires normals with consistent orientation or inside/outside information, which is not trivial to be obtained from raw point clouds. The representative one is Poisson Surface Reconstruction~\cite{2006Poisson, 2013Screened, 2020EnvelopePoisson}, which establishes Poisson equations to solve the indicator function whose gradient is close to the vector field. Calakli et al.~\cite{2011SSD} apply optimization methods to approximate a smooth signed distance function of the surface. Fuhrmann et al.~\cite{2014FloatingScale} assign a compactly supported basis function to each point and constructs the implicit function through the sum of the basic functions. GR~\cite{2019GaussRecon} is a non-data-driven method also motivated by Gauss lemma. It obtains the indicator value at the query point through the surface integral of the modified kernel function. VIPSS~\cite{2019VIPSS} is a recent classical reconstruction method that does not require normals and achieves good results in sparse and sketch samplings. However, the time complexity of this method is proportional to the cubic of the point number. The process may take a few days even if the point number is about twenty or thirty thousand.

\subsection{Data-driven reconstruction}
With the rapid development of deep learning in 3D geometry~\cite{2017PointNet, 2017PointNet++}, researchers have become increasingly interested in data-driven approaches. Mescheder et al.~\cite{2019Occupancy} implicitly represent 3D surface as the continuous decision boundary of a neural network classifier. Peng et al.~\cite{2020ConvOccupancy} improve this work by combining the strengths of convolutional neural networks with implicit representations. DeepSDF~\cite{2019DeepSDF} learns the latent spaces of shapes. It compiles the entire shape into a single latent vector and regresses the continuous signed distance function from the latent codes. However, this approach is data-specific and cannot adapt to generalize reconstruction problems. Chabra et al.~\cite{2020DeepLocalShapes} use local shape latent vectors to decompose the overall complex shape into simple local shapes. However, it lacks global information. Thus, it can not infer a global and object-level latent description. Atzmon et al.~\cite{SAL2020} and Zhao et al.~\cite{SALRecon2020} propose sign agnostic learning approaches for surface reconstruction, which push a network to fit a signed version of the unsigned function when given proper initialization of network weights. However, they need to operate a reinitialized network for each shape. Points2Surf~\cite{2020Points2Surf} is the work most relevant to us. It learns the absolute distance through the local patch and obtains the inside/outside information from the global sampling. However, it is not good at normal consistency and may produce bumpy surfaces with ripples.

\section{Preliminary} \label{Preliminary}

Our network is designed mainly based on Gauss lemma in potential energy theory~\cite{1995PDE, 2009Potential}. We first introduce Gauss lemma in this section and identify how we modify the traditional indicator function for better training.

\emph{Gauss Lemma: If $\Omega$ is an open and bounded set with a smooth boundary in $\mathbb{R}^3$. For any $x \in \mathbb{R}^3$, $y \in \partial \Omega$, let $\chi(x)$ be the following surface integral of $\partial \Omega$:}
\begin{equation}
\label{equation1}
\chi(x) = \int_{\partial \Omega} \frac{\partial G}{\partial \vec{N}(y)}(x, y)dS(y),
\end{equation}
\emph{where $\vec{N}(y)$ denotes the outward unit normal vector at $y$,  $dS(y)$ represents the surface element and $G$ is the fundamental solution of Laplacian equation in $\mathbb{R}^3$. That is, for $x, y \in \mathbb{R}^3$, }
\begin{equation}
\label{equation2}
G(x, y) = - \frac{1}{4\pi ||x-y||}.
\end{equation}
\emph{Then, the result of the integral $\chi(x)$ is exactly the indicator function}
\begin{equation}
\label{equation3}
\chi(x)=\left\{
\begin{aligned}
&0 & & {x \in \mathbb{R}^3  \backslash \overline{\Omega}} \\
&\frac{1}{2}  & &   {x \in \partial \Omega} \\
&1 & & {x \in \Omega}
\end{aligned} \right.
\end{equation}
The deriviate in Equation~\ref{equation1} has the following expression:
\begin{equation}
\label{equation4}
\frac{\partial G}{\partial \vec{N}(y)}(x, y) = - \frac{1}{4\pi} \frac{(x-y) \cdot \vec{N}(y)}{{||x-y||}^{3}}.
\end{equation}
The above indicator function is discontinuous.
Thus, we modify the indicator function and smoothen it near the surface. The ideal function is linear near the surface, while keeping the ordinary zero-one function away from the surface. Specifically,
\begin{equation}
\label{equation5}
\tilde{\chi}(x)=\left\{
\begin{aligned}
&0 & & {d(x, \Omega) < -w} \\
&\frac{1}{2} + \frac{d(x, \Omega)}{2w} & & {|d(x, \Omega)| \leq w} \\
&1 & & {d(x, \Omega) > w} \\
\end{aligned} \right.
\end{equation}
where $d(x, \Omega)$ is the signed distance of $x$ to $ \Omega$ (positive inside and negative outside), $w$ is a small and fixed width coefficient. We take $\tilde{\chi}(x)$ as the objective function to learn.

\begin{figure}[htb]
  \centering
  \includegraphics[width=1.0\linewidth]{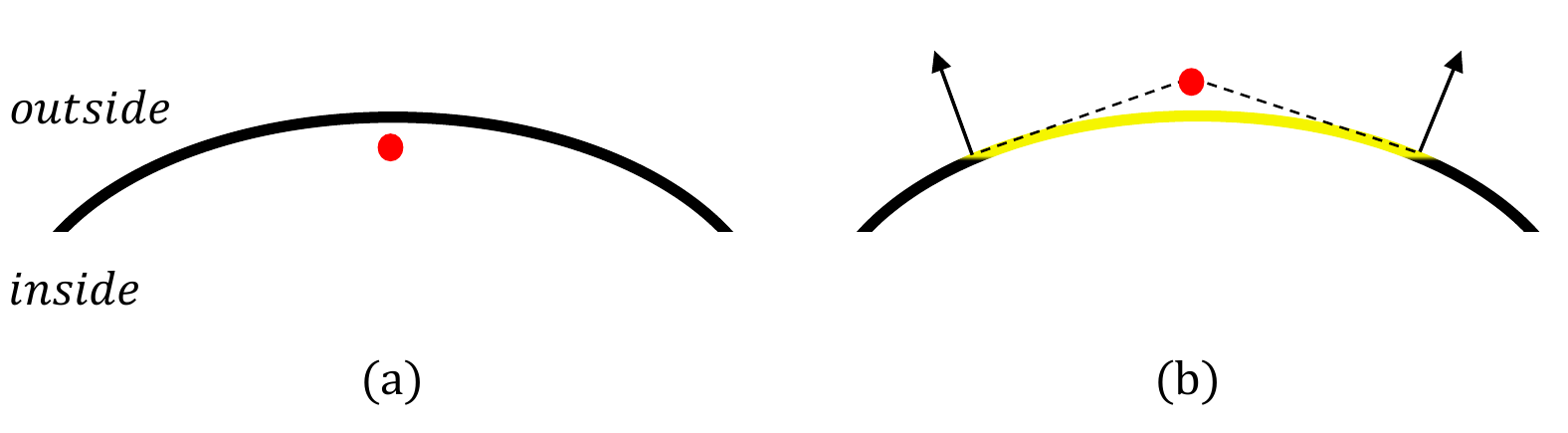}
  \caption{
            Contributions to the integral. We use different colors to distinguish between positive and negative contributions. The black curve represents positive contributions, while the yellow curve represents negative contributions. The red ball represents the query point.}
    \label{fig:Figure3}
\end{figure}

\begin{figure*}[htb]
  \centering
  \includegraphics[width=1.0\linewidth]{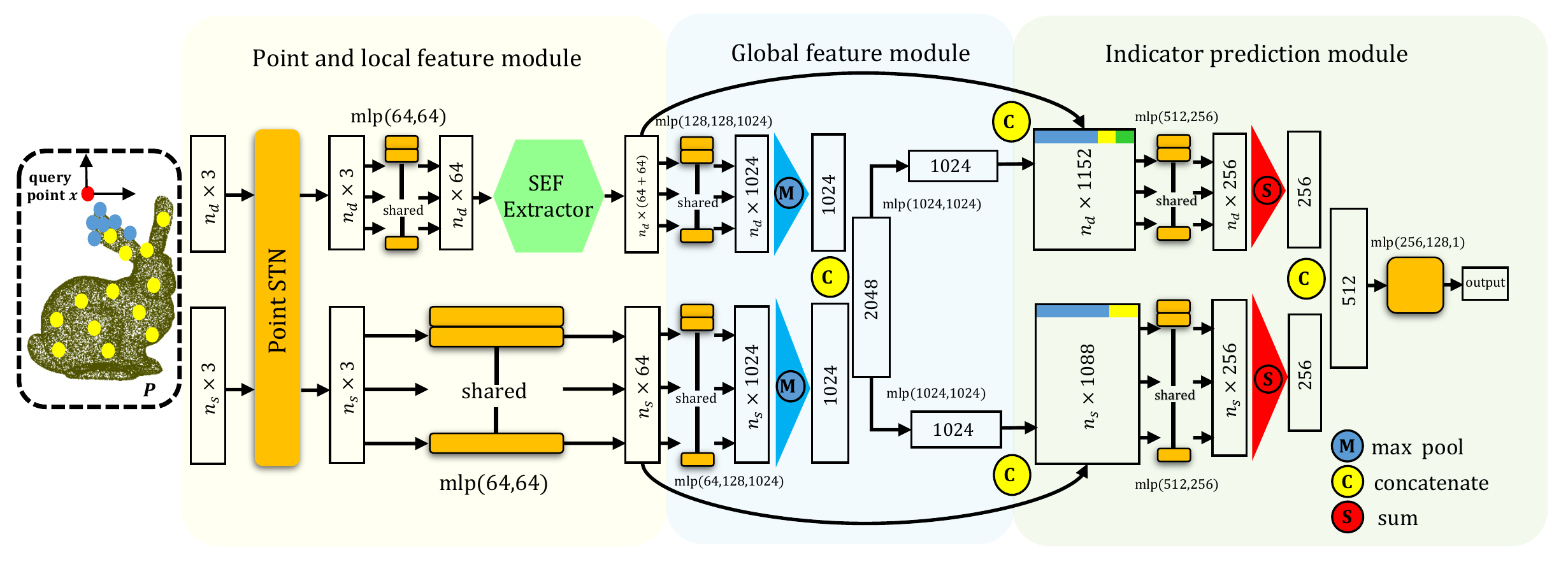}
  \caption{
           Our network architecture. The network has two parallel branches horizontally and consists of three main modules vertically. To apply Gauss lemma in our network, we propose a novel Surface Element Feature Extractor (SEF Extractor) for learning the local shape properties and we sum the point-wise contributions to process the surface integral in the network.}
    \label{fig:Figure4}
\end{figure*}

\section{Method} \label{Method}
\subsection{Motivations} \label{MethodMotivations}

Given a 3D point cloud $P=\{p_1, p_2,...,p_n \}$, our goal is to reconstruct a watertight surface $S$ to express its shape. Therefore, we aim to train a network that can predict the modified indicator function $\tilde{\chi}_S(x)$ for point clouds from arbitrary shapes. 

Gauss lemma indicates that for a query point $x \in \mathbb{R}^3$, the indicator value of $x$ can be directly obtained by surface integral. If $x$ is near the surface and moves from inside to outside, as shown in Fig.~\ref{fig:Figure3}, the indicator value changes from one to zero. The change is mainly contributed by the curve close to $x$. If we use different colors to distinguish between positive and negative contributions to the integral, we can notice that the contributions of the yellow curve on Fig.~\ref{fig:Figure3}(b) change from positive to negative when the query point moves from inside to outside. This mainly causes the changes in the indicator value. This indicates that local surface properties account for the major contributions to the final result. Therefore, we need a dense sampling near the query point. 

Similar to Points2Surf~\cite{2020Points2Surf}, for each query point $x$, we take a local patch $p_{x}^{d}$ and a global sub-sample $p_x^{s}$ from the original point cloud $P$. The set $p_x^{d}$ is composed of the $n_{d}$-nearest neighbors of $x$ ($n_{d} = 200$ in our experiments). $p_x^{s}$ is obtained by uniformly sampling $n_{s}$ points from the entire point cloud ($n_{s} = 1000$ in our experiments). The dashed box in Fig.~\ref{fig:Figure4} visualizes our sampling approach. The blue balls represent the local patch and the yellow balls represent the global sub-sample. We center $p_x^{d}$ and $p_x^{s}$ at the query point. Thus, we do not need to input the query point into the network explicitly. Then, our model can be expressed as a function $f_{\theta}: \mathbb{R}^{3n_{d}} \times \mathbb{R}^{3n_{s}} \rightarrow \mathbb{R}$. We are looking to train the parameters $\theta$ of the network, such that, given any sample $p_x^{d}$ and $p_x^{s}$, $f_{\theta}(p_x^{d}, p_x^{s})$ can better approximate $\tilde{\chi}_S(x)$.

Let $\mathcal{Y} = \{y_{i}\}_{i=1}^{N} \subset \partial \Omega$ be a sample point cloud of the surface. To perform surface integral in Gauss Lemma, we write Equation~\ref{equation1} in the form of point-wise contributions. 
\begin{equation}
\label{equation6}
\chi(x) = \sum_{i=1}^{N}{ \underline{-\frac{1}{4\pi} \frac{(x-y_i)}{{||x-y_i||}^{3}}}_{1} \cdot \underline{\vec{N}(y_i)}_{2} \cdot \underline{\sigma(y_i) }_{3}}.
\end{equation}

The first part is only related to the query point $x$ and sample point $y_i$. Thus, it can be expressed by point features. The third part approximates the surface element and is related to local properties. Hence, we need to learn local features from the nearest neighbors. The second part is about the normals. The estimation of the unsigned normals only needs local properties, whereas the estimation of the signs still requires global information. We can learn a global latent vector to compile and summary the entire shape. We aggregate features with different scales from local to global to obtain accurate point-wise contributions. In this way, we implicitly apply Gauss lemma in the neural network.

\subsection{Overview} \label{MethodOverview}

The overall network architecture is visualized in Fig.~\ref{fig:Figure4}.
Horizontally, the network has two parallel branches, corresponding to patch feature branch and shape feature branch, respectively. Shape feature branch takes the global sub-sample as input. It learns the overall shape properties and makes a rough estimation for the indicator value. Patch feature branch takes the local patch sample as input, compiling local information for accurate indicator prediction.

Vertically, the network has three main modules, namely, \textit{point and local feature module}, \textit{global feature module}, and \textit{indicator prediction module}. They correspond to three different consecutive stages. \textit{Point and local feature module} mainly learns the local properties of each point. In this module, we propose a novel \textbf{Surface Element Feature Extractor} (SEF Extractor), which aggregates the point features itself with the features of its spatial $k$-nearest neighbors in the sampling set $(p_x^{d}, p_x^{s})$ and obtains the local properties through a miniature symmetric network. This extractor also corresponds to the surface element $dS(y)$ and the normal $\vec{N}(y)$ in Gauss lemma (also $\sigma(y_i)$ and $\vec{N}(y_i)$ in Equation~\ref{equation6}). The specific details of this module will be discussed in Section~\ref{localfeature}. \textit{Global feature module} mainly compiles the global latent vector. The specific details of this module will be discussed in Section~\ref{globalfeature}. In the\textit{ indicator prediction module}, we concatenate features with different scales and perform surface integral throughout the sum of the point-wise contributions in the network. The final prediction will be given in this module. The specific details of this module will be discussed in Section~\ref{indicator_prediciton}.

\subsection{Point and local feature module} \label{localfeature}
\begin{figure}[htb]
  \centering
  \includegraphics[width=1.05\linewidth]{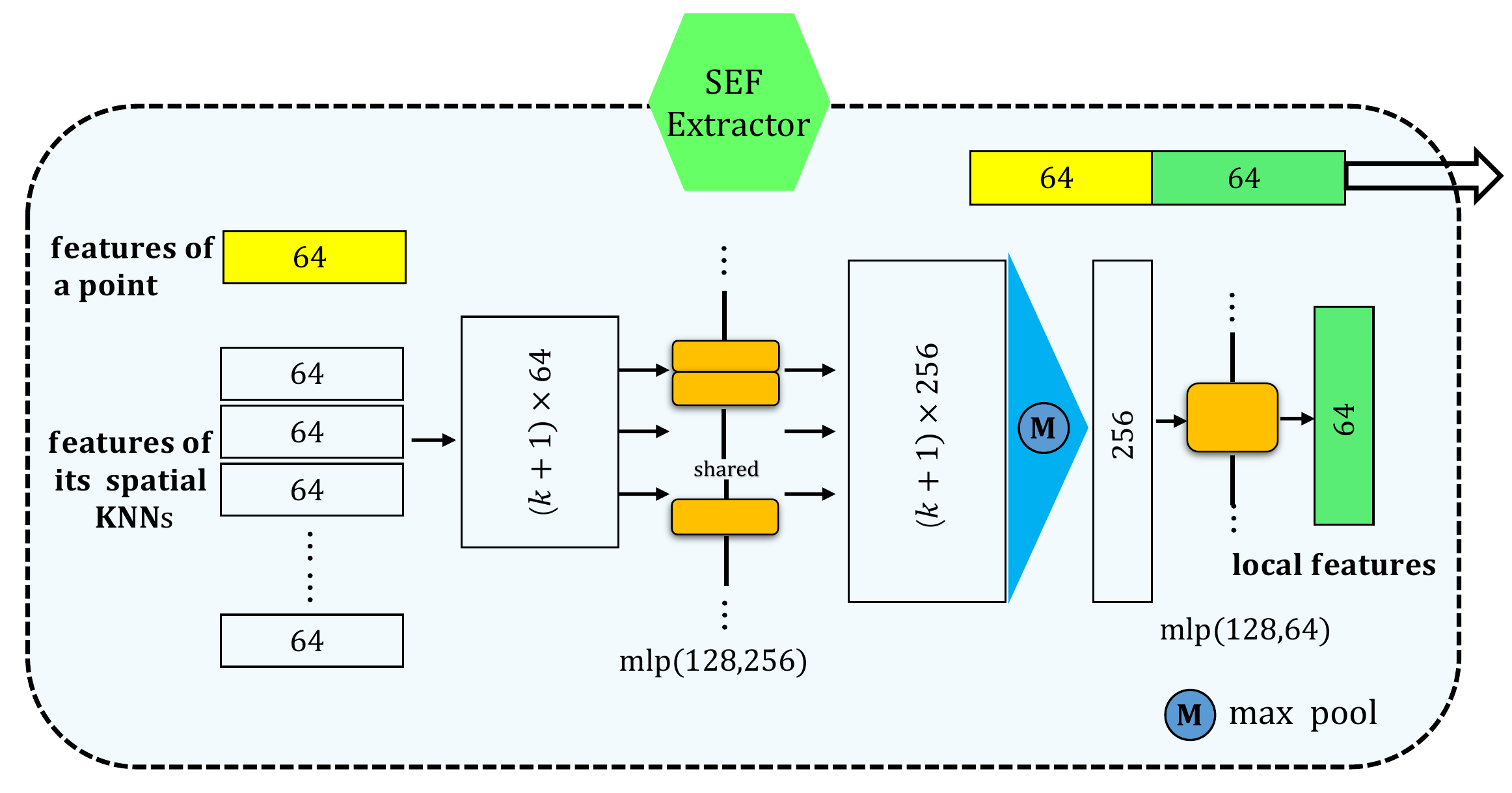}
  \caption{
           Illustration of the Surface Element Feature Extractor (SEF Extractor). It is a symmetric network. For each point, the extractor aggregates the point features itself, and the features of its spatial $k$-nearest neighbors (KNNs) to compile the local features. We concatenate the point features and the local features for output.}
    \label{fig:Figure5}
\end{figure}

Our goal in this module is to extract point and local features. In particular, we need accurate local shape information for points close to $x$. Notice that we center all the sub-samples to $x$, so the indicator value is rotation invariant corresponding to the coordinate origin. Thus, we apply a point Spatial Transformer Networks (STN)~\cite{STN15} to all the sub-samples. Several multilayer perceptrons (MLPs) are applied to obtain the 64-dimensional point features. However, point features are insufficient to compile the local properties. Local properties usually include the local shape details, sampling density, and local noise scales. Only with these information can we learn the normals and the surface elements. Thus, we propose a novel SEF Extractor to aggregate local information. This name comes from the surface element $dS(y)$ in Gauss lemma. The detailed architecture of the SEF Extractor is presented in Fig.~\ref{fig:Figure5}.

\begin{figure*}[htb]
  \centering
  \includegraphics[width=1.0\linewidth]{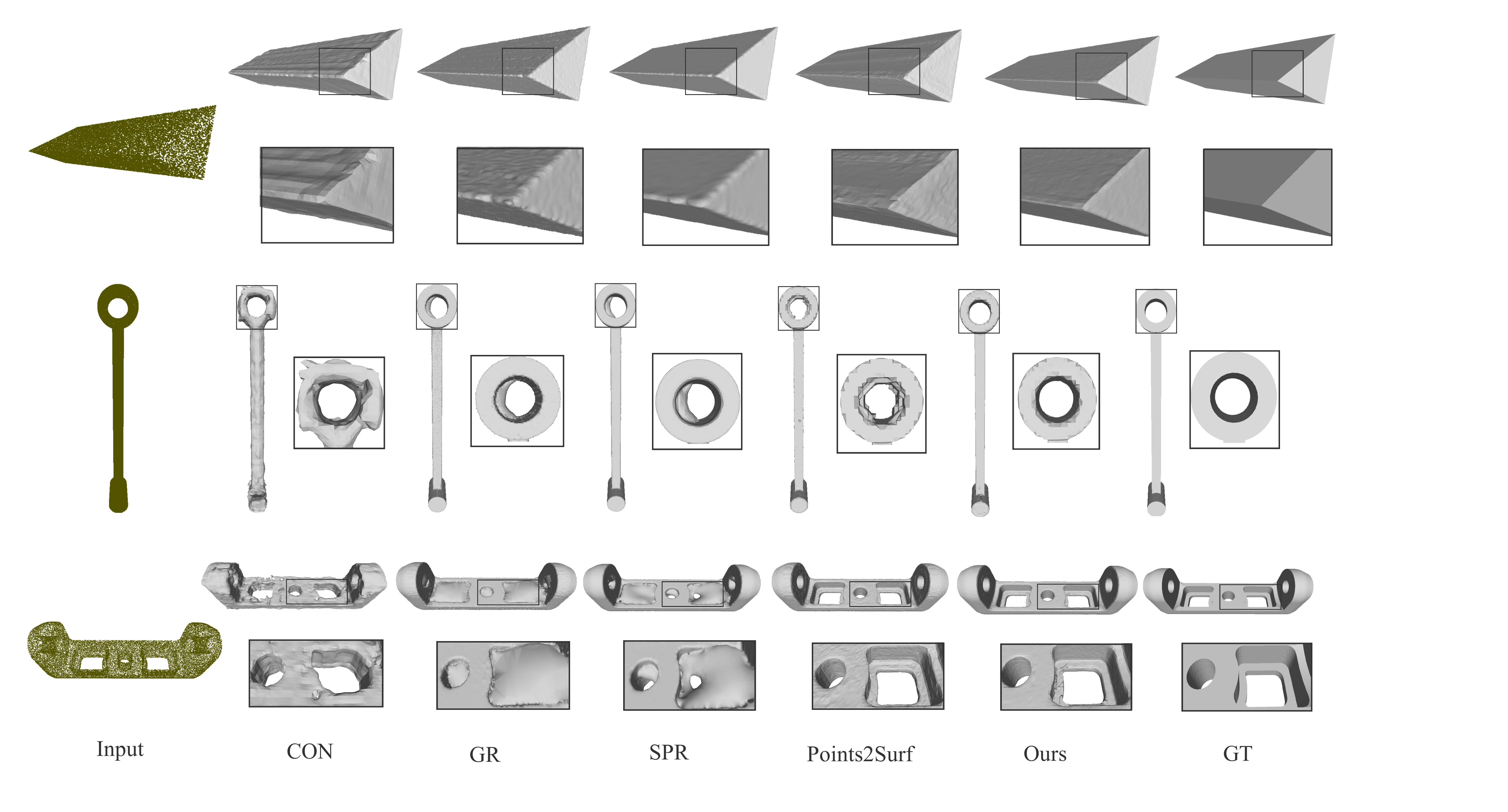}
  \caption{\label{fig:Figure6}
           Qualitative comparison on the ABC no-noise dataset.}
\end{figure*}

\begin{figure*}[htb]
  \centering
  \includegraphics[width=1.0\linewidth]{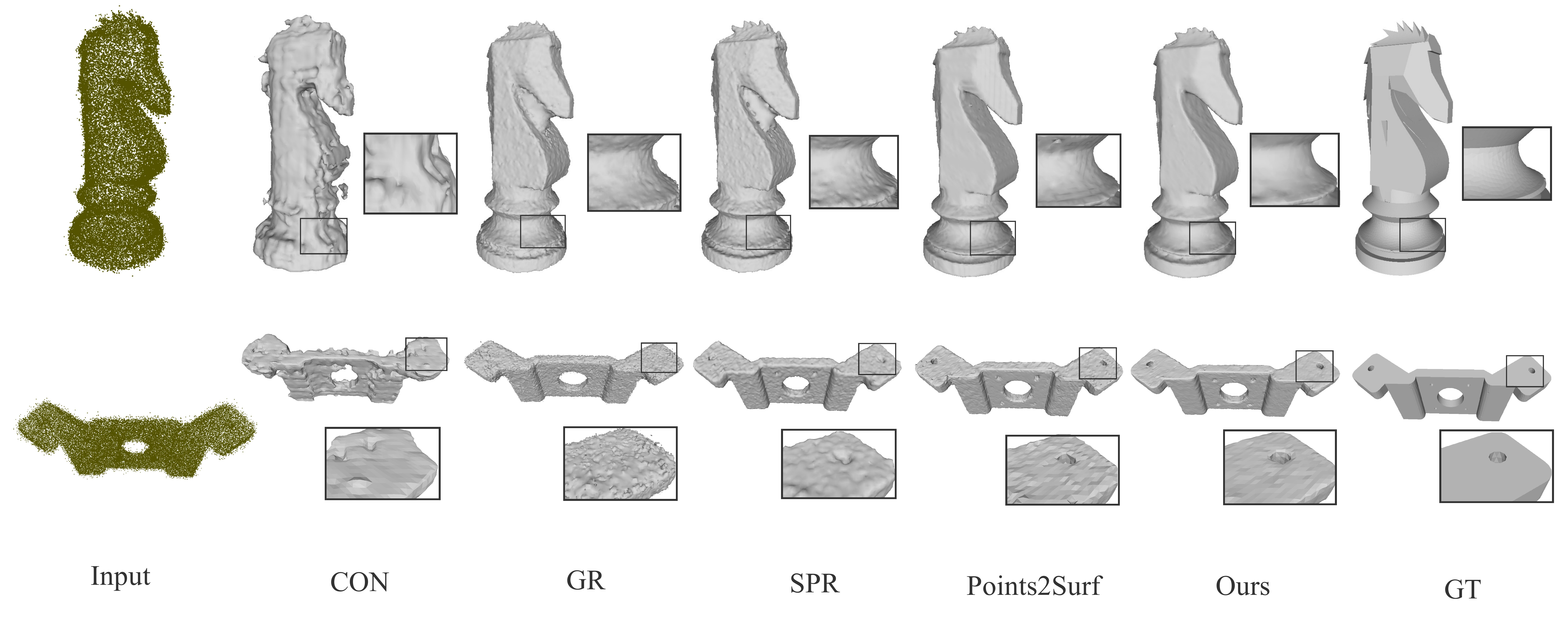}
  \caption{\label{fig:Figure7}
           Qualitative comparison on the ABC var-noise dataset.}
\end{figure*}

\begin{figure*}[htb]
  \centering
  \includegraphics[width=1.0\linewidth]{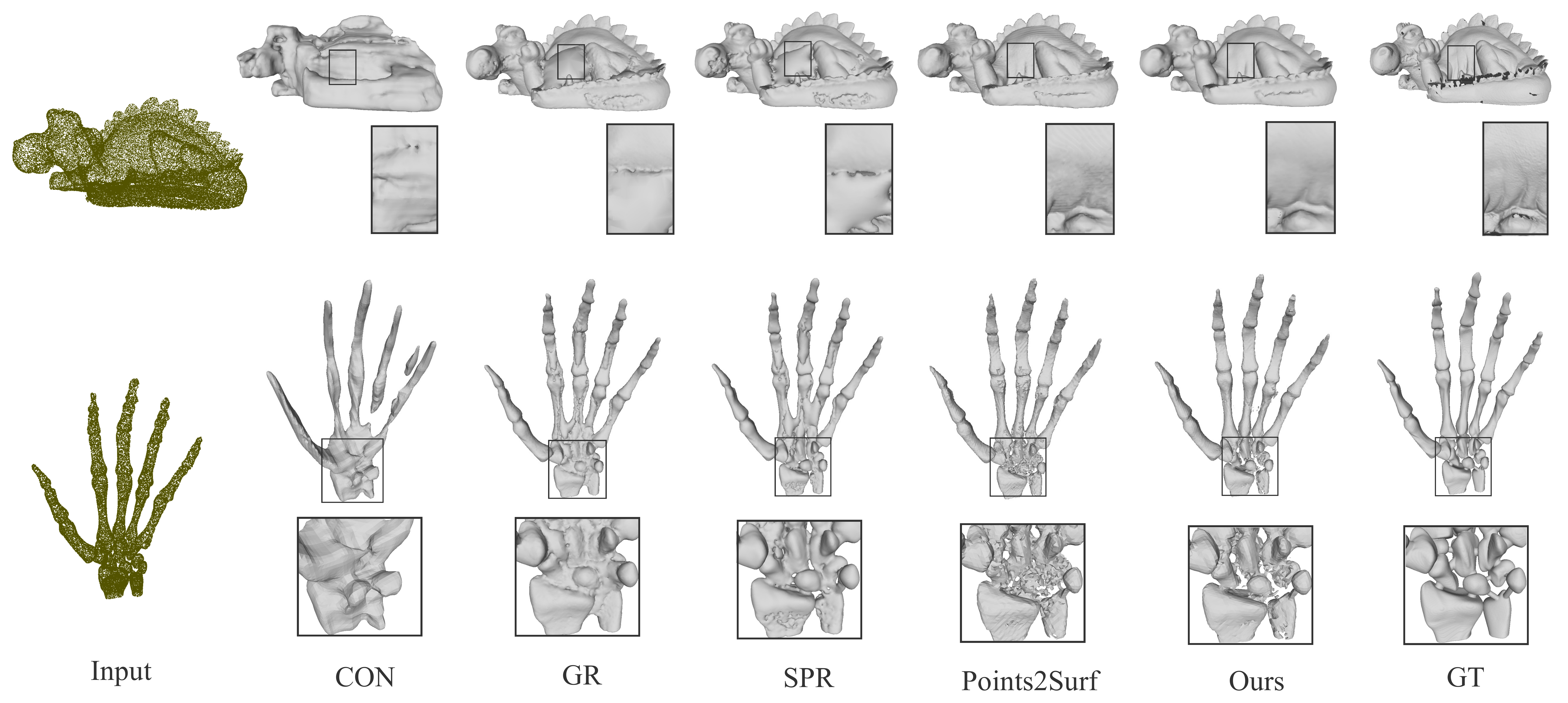}
  \caption{\label{fig:Figure8}
           Qualitative comparison on the Famous no-noise dataset.}
\end{figure*}

\begin{figure*}[htb]
  \centering
  \includegraphics[width=1.0\linewidth]{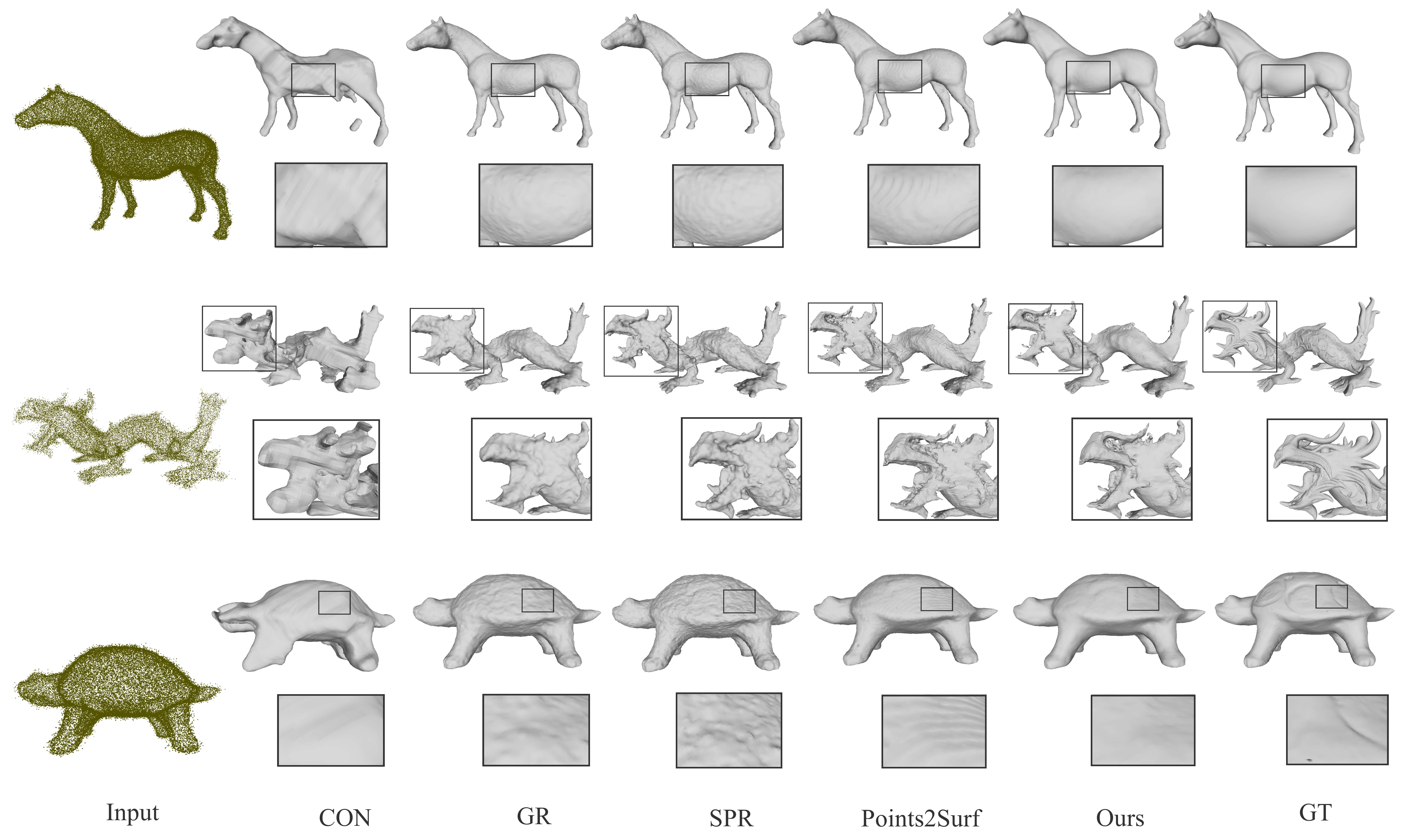}
  \caption{\label{fig:Figure9}
           Qualitative comparison on the Famous var-noise dataset.}
\end{figure*}

\begin{figure*}[htb]
  \centering
  \includegraphics[width=1.0\linewidth]{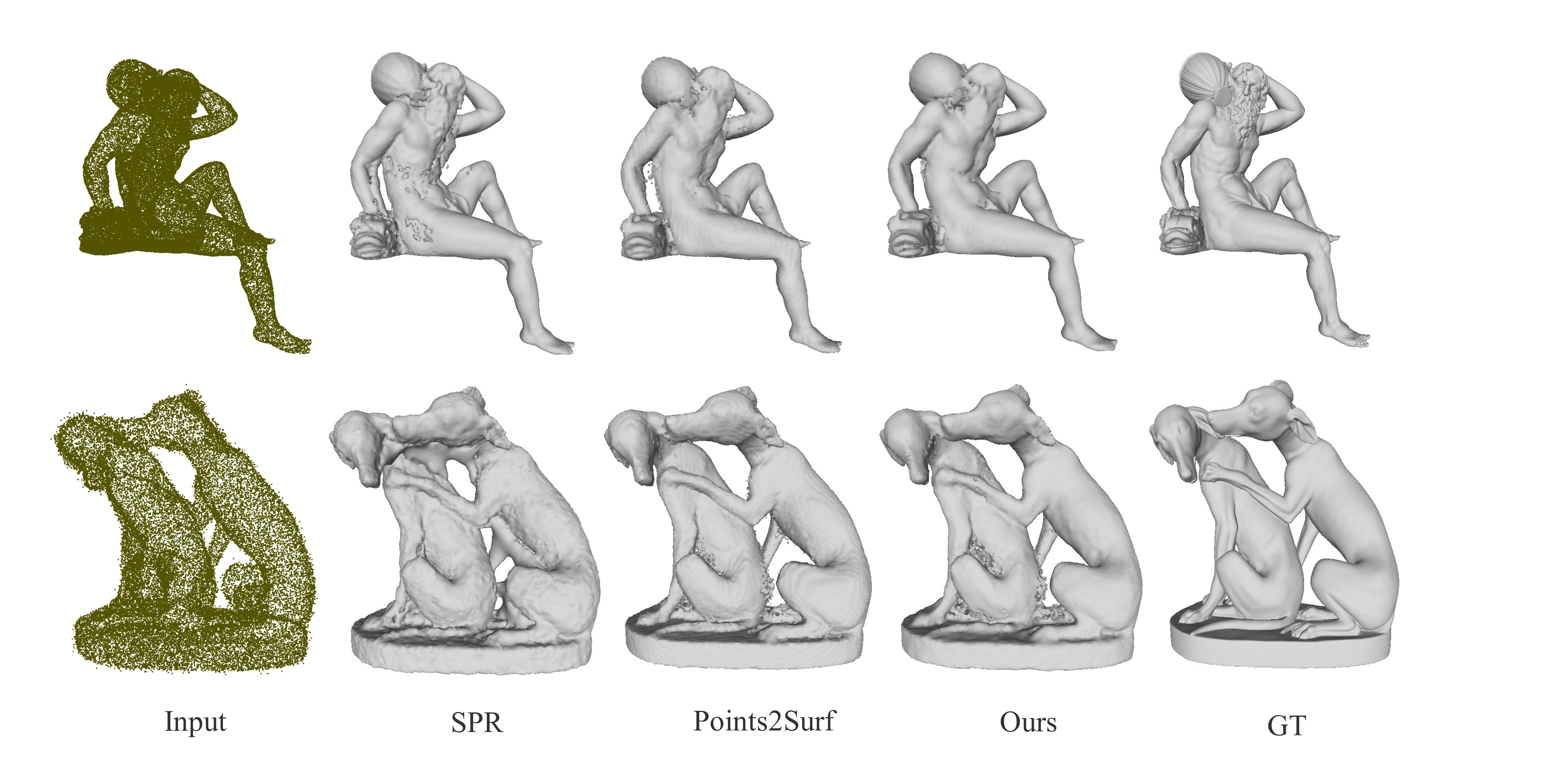}
  \caption{\label{fig:Figure10}
           Qualitative comparison on the ThreeDScans dataset. The first example is a clean sampling and the second example is a noisy sampling.}
\end{figure*}

\begin{figure}[htb]
  \centering
  \includegraphics[width=1.0\linewidth]{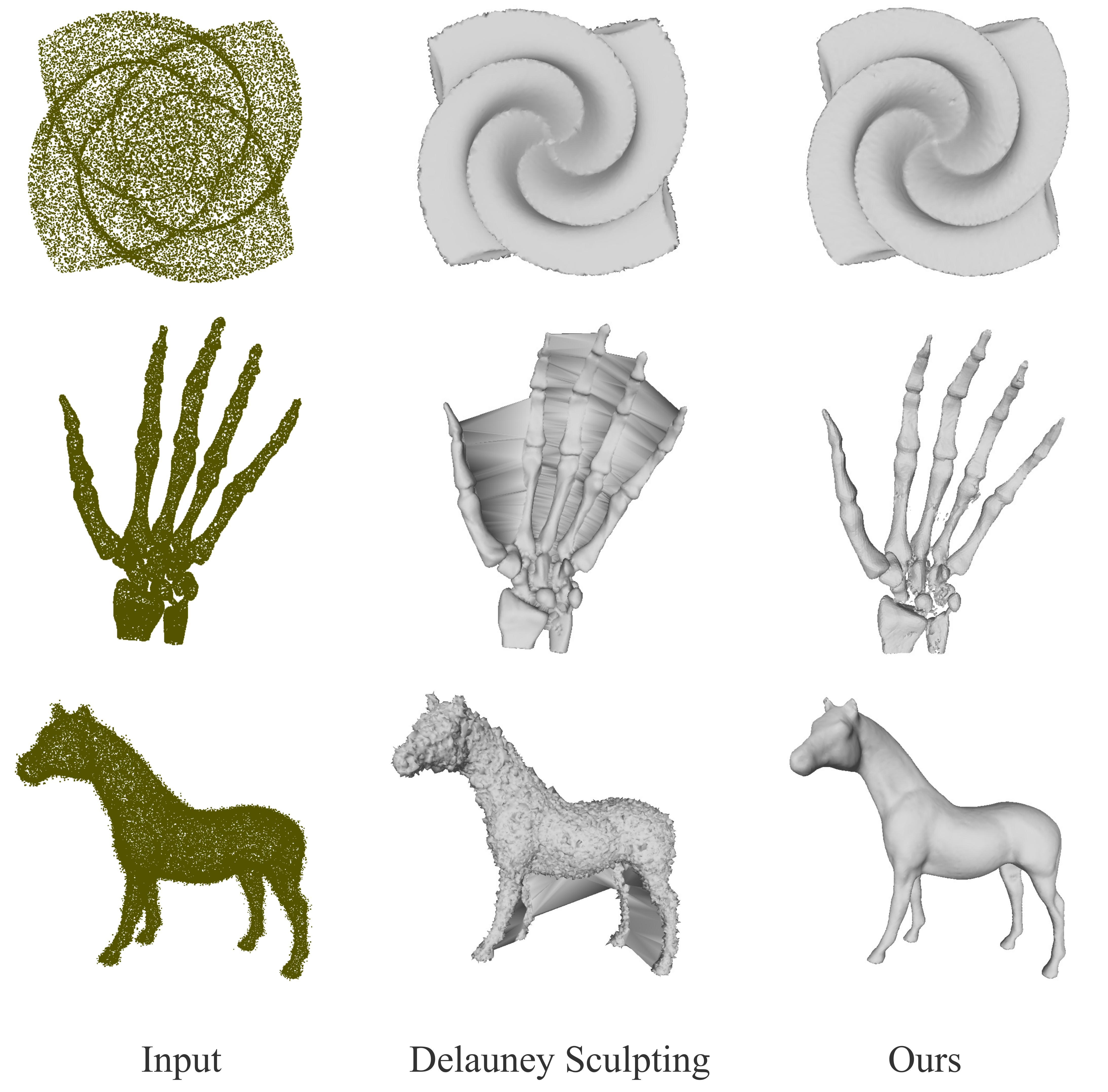}
  \caption{\label{fig:Figure11}
           Qualitative comparison with explicit reconstruction method Delauney Sculpting ~\cite{2015Delaunay}. The first and the second examples are clean samplings and the third example is a noisy sampling.}
\end{figure}

\begin{figure*}[htb]
  \centering
  \includegraphics[width=1.0\linewidth]{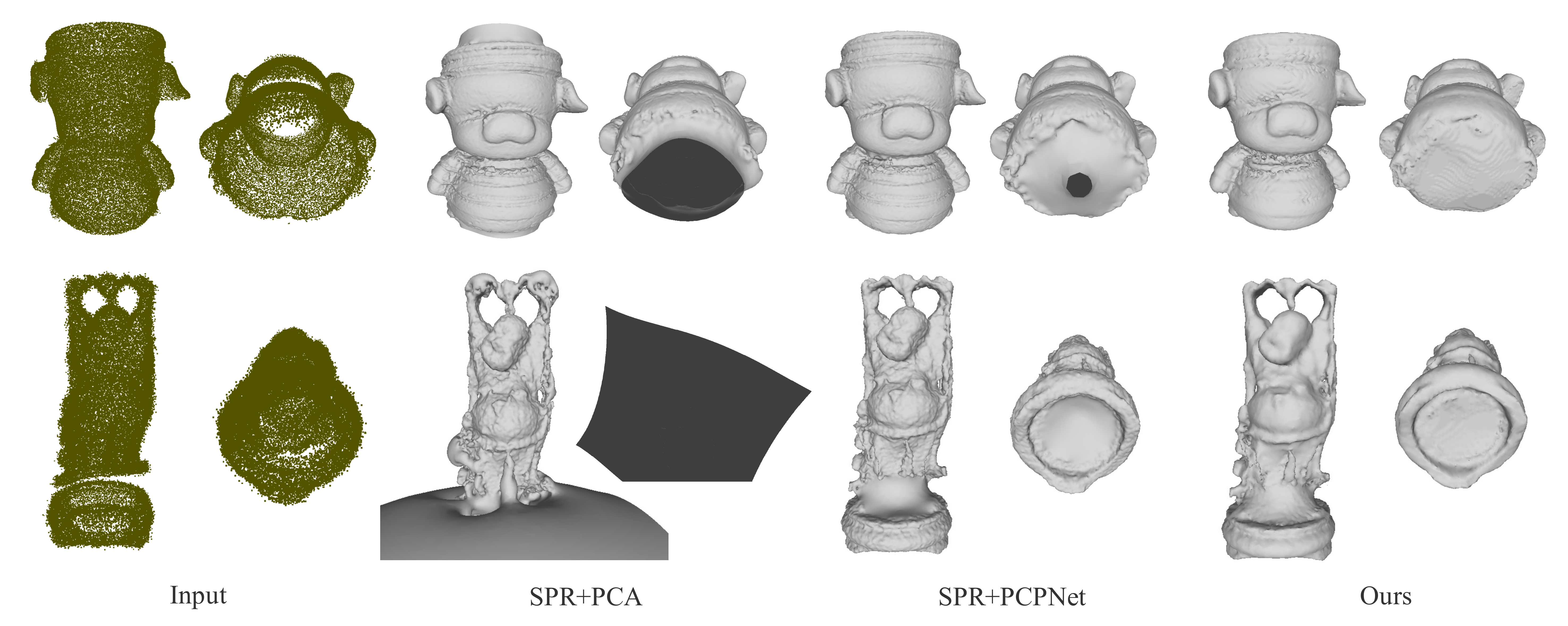}
  \caption{\label{fig:Figure12}
           Qualitative comparison on the data with missing parts. Our method can effectively close the holes caused by missing parts.}
\end{figure*}

\begin{figure}[htb]
  \centering
  \includegraphics[width=1.0\linewidth]{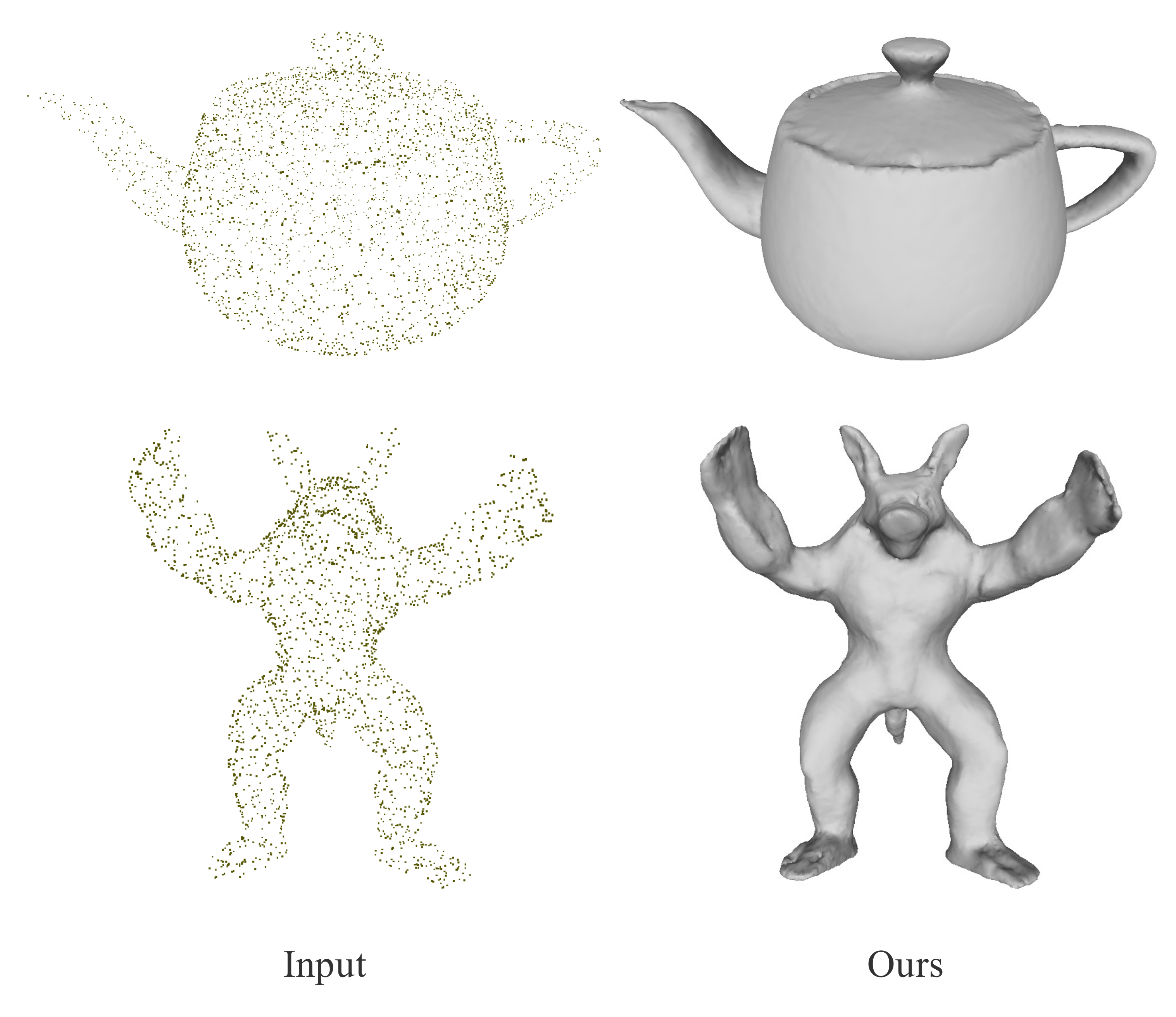}
  \caption{\label{fig:Figure13}
           Our reconstructions of sparsely sampled point clouds.}
\end{figure}

The key idea of the SEF Extractor is to learn local shape properties via the spatial $k$-nearest neighbors (KNNs). For each point, we concatenate the features of itself and the features of its spatial KNNs in the sampling set $(p_x^{d}, p_x^{s})$. They are fed into a miniature symmetric network consisting of several MLPs and a symmetric max pooling operation to learn a $64$-dimensional vector as the local features for each point. Then, we concatenate the local features with the point features and finally output a $128$-dimensional vector. This vector contains sufficient local information.  From Equation~\ref{equation4}, we notice that $\frac{\partial G}{\partial \vec{N}(y)}$ drops quickly when $x$ moves away from $y$. Thus, we only apply the SEF Extractor to the patch feature branch, which usually accounts for the major contributions to the final indicator prediction. The parameters of the MLPs are shared among all the $n_d$ points.

During implementation, gathering KNN features in each forward and backward process in the neural network may be extremely time-consuming. However, considering that the parameters of MLPs in the SEF Extractor are shared among all points, gathering the KNN features in forward and backward processes is equivalent to concatenating their coordinates directly before inputting them into the network. Therefore, we are only required to perform $n_d$ searches on a k-d tree with $n_s + n_d$ leaf nodes for each sampling. Given that these operations can be executed in parallel with forward and backward processes in the neural network, they do not consume considerable additional time. The time for finishing the same amount of data training is about 1.3 times that of Points2Surf. We use $k = 10$ in all the experiments and conduct ablation studies for other values in section ~\ref{ablation_study}.

\subsection{Global feature module} \label{globalfeature}
We aim to learn the latent description of shapes in this module. We aggregate point features by max pooling as PointNet~\cite{2017PointNet} and obtains 1024-dimensional patch and shape features. Then, we concatenate them to obtain the 2048-dimensional global features. The latent vector is computed by applying several fully connected layers to the global features. The parameters of 
the fully connected layers are different for patch feature branch and shape feature branch. We will further concatenate the global latent vector with local features for accurate point-wise contributions and indicator value predictions. 

Note that outward normals are required for the Gauss Lemma. We need the global latent vector to determine inward or outward. Qi et al.~\cite{2017PointNet} indicate that the global latent vector tends to summarize an entire shape with a sparse set of key points. This description can be used to explain why the global latent vector can be applied to determine the orientation of normals and the signs of indicator values. The idea of shape summarizing is also reflected in some classical methods. For example, Huang et al.~\cite{2009Consolidation} create thinned and evenly distributed sets of points for oriented normal estimation and point consolidation. 

\subsection{Indicator prediction module} \label{indicator_prediciton}

In this module, we concatenate the features of different scales from local to global latent for each point. In the patch feature branch, we concatenate the point features ($64$ dim), local features ($64$ dim), and the global latent vector ($1024$ dim). In this manner, we comprehensively aggregate properties with different scales, such as local shape details, sampling density, noise scales, and orientation information. Referring to Equation~\ref{equation6}, we obtain sufficient information for accurate point-wise contributions. For the shape feature branch, we only concatenate the point features and the global latent vector. We perform the surface integral by aggregating the point-wise contributions in the neural network. In the shape feature branch, the indicator value is roughly estimated. And in the patch features branch, we further enhance and refine the final result. Therefore, we still need several fully connected layers to combine patch and shape features before obtaining the final predictions.

Points2Surf~\cite{2020Points2Surf} directly predicts the signed distance at a query point from the latent codes. By contrast, we reuse the latent vector for accurate point-wise information. We in turn use the orientation of each point to supervise the global latent codes. Thus, our method can effectively learn the latent vector. We also distribute potential errors to sample points. Thus, the surface generated by our approach is smooth and shows good normal consistency. The SEF Extractor we proposed is also helpful for distinguishing noise.

Our network is end-to-end. During training, we directly apply L$2$ regression to the aforementioned modified indicator function $\tilde{\chi}(x)$ in Equation~\ref{equation5}. This is simple and effective. The surfaces are generated by applying Marching Cubes algorithm~\cite{1987MarchingCubes, 2013EfficientMarching} to the modified indicator functions.

\section{Experiments} \label{experiments}
\subsection{Overview} \label{experiments_overview}
We conduct plenty of experiments on various datasets with different noise scales to validate the efficacy of our method in terms of accuracy, geometric details, and generalizability. We also examine the ability of our method to manage artifacts and very sparse samplings. We first present an overview of our experiments.

\underline{\textbf{Datasets}}

\textbf{ABC}~\cite{2019ABC}: The ABC dataset is a big collection of Computer-Aided Design (CAD) models for geometric deep learning. The models exhibit high quality with accurate normals. We use this dataset for training and evaluation.

\textbf{Famous}: This dataset contains $22$ well-known shapes in geometry processing, such as the Stanford Bunny and the Utah teapot. Erler et al.~\cite{2020Points2Surf} compiled these shapes together and called them Famous dataset. We only use this dataset for testing, i.e., no training data is selected from this dataset. Thus, we primarily use this dataset to examine the generalizability of different methods.

\textbf{ThreeDScans}~\cite{3DScansWeb}: This dataset contains high-quality lifelike sculptures. We use it to examine the capability of different methods to construct objects with complex geometric topology. This dataset is also used only for testing.

\underline{\textbf{Baselines}}

\textbf{SPR}~\cite{2013Screened}: Screened Poisson Surface Reconstruction (SPR) is regarded as the gold standard for non-data-driven surface reconstruction. It requires oriented normals for input. Thus, we utilize the recent PCPNet~\cite{2018PCPNet} with multi-scale architecture to estimate normals directly from raw point clouds due to its robust to noise.

\textbf{GR}~\cite{2019GaussRecon}: GR is a non-data-driven reconstruction method based on the modified Gauss formula. It also requires oriented normals with global consistency.

\textbf{CON}~\cite{2020ConvOccupancy}: Convolutional Occupancy Networks (CON) is a learning-based method that combines convolutional encoders with implicit occupancy decoders. It can learn the occupancies directly from raw point clouds without normals.

\textbf{Points2Surf (P2S)}~\cite{2020Points2Surf}: Points2Surf (P2S) is a representative learning-based approach for general surface reconstruction (not data-specific and generalized to various shapes). We use it as a critical baseline. It combines a detailed local patch and a coarse global sampling to learn the signed distance function directly for arbitrary point clouds without normals. 

\underline{\textbf{Metrics}}

\textbf{Two-way chamfer distance (CD)}: Two-way chamfer distance (CD) is a simple and popular metric that sums the nearest-neighbor distance between two densely sampled point sets for each mesh. The formula for calculating CD is given as follow:

\begin{small}
\begin{equation}
\label{equation8}
CD(A, B) = \frac{1}{|A|} \sum \limits_{p_i \in A} \min \limits_{p_j \in B} {||p_i - p_j||_{2}} 
+ \frac{1}{|B|} \sum \limits_{p_j \in B} \min \limits_{p_i \in A} {||p_j - p_i||_{2}}.
\end{equation}
\end{small}

\textbf{Normal consistency error (NCE)}: Chamfer distance is insufficient for estimating the smoothness and accuracy of mesh normals. Hence, we also estimate the normal consistency error (NCE) between the reconstructed and ground truth meshes. Normal consistency is the mean cosine similarity between the normals of the sampled points from the ground truth mesh and the nearest face on the reconstructed mesh. NCE is defined as one minus normal consistency. The formula for calculating NCE is given as follow:
\begin{equation}
\label{equation9}
NCE(A, GT) = 1 - \frac{1}{|GT|} \sum \limits_{(p_i, n_i) \in GT} {|n_i \cdot n_j|},
\end{equation}
where $n_j$ is the normal of the face closest to $p_i$ in $A$.

\underline{\textbf{Data preparation}}

For the ABC dataset, we randomly select $7495$ watertight meshes for training and $100$ for testing. The richness of training shapes is critical for accurate global feature learning. We normalize each mesh, scaling its longest length to $1.0$ and randomly sampling $20,000$ to $80,000$ points on the surface. The sampling density can also be flexibly selected according to needs.

To ensure that our method can adapt to different noise scales, and simultaneously, obtain a smooth and consistent mesh under a relatively clean sample, we randomly select $10\%$ perfect shapes without noise. For other shapes, we randomize a noise ratio $\alpha_P$ for each shape. That is, the ratio with noise is $\alpha_P$ and the ratio without noise is $1 - \alpha_P$ for all sample points of this shape. We also randomize a noise amplitude within $[0.02, 0.04]$ for each shape and add randomized Gaussian noise for each noisy point. Here, an amplitude $\beta$ refers to a point-wise Gaussian noise with a maximum value of $\beta$ and a standard deviation of $\frac{\beta}{3}$. For each shape, we randomly sample $800$ query points near the surface and $200$ in the entire unit cube for training. Then, we have 1000 different training samples $(p_x^{d}, p_x^{s})$ for each shape and up to $7 \times 10^{6}$ training samples with varying noise in total. We use the trimesh library~\cite{2019trimesh} to calculate the signed distance functions (SDFs). Then, the ground-truth indicator functions can be obtained by putting SDFs into Equation~\ref{equation5}. We set the grid size to $\frac{1}{256}$ and the width coefficient $w$ to $4$ times the grid size in Equation~\ref{equation5}.

\subsection{Results} \label{results}

\begin{table}[t]
\centering
\caption{Quantitative comparison on the ABC test dataset. We multiply the CD value by 100. Our method achieves the best results on both clean data and noisy data. Among the 100 test shapes, our method exhibits the best normal consistency among all the methods in $63$ and $80$ shapes respectively.}
\label{table1}
\begin{tabular}{ccccccc}
\hline
\multirow{2}{*}{Method} & \multicolumn{3}{c}{ABC no-noise} & \multicolumn{3}{c}{ABC var-noise}\cr
\cmidrule(lr){2-4} \cmidrule(lr){5-7}
  & CD & NCE & BCR & CD & NCE & BCR \\
\hline

CON  & $1.742$ & $0.102$ & 2\% & $1.627$ & $0.094$ & 6\%\\

GR  & $1.703$ & $0.109$ & 0\% & $1.707$ & $0.135$ & 0\% \\

SPR  & $1.588$ & $0.052$ & 33\% & $1.633$ & $0.070$ & 10\% \\

P2S  & $1.126$  & $0.048$ & 2\% & $1.132$ & $0.056$ & 4\% \\

Ours  & $\textbf{1.110}$  & $\textbf{0.042}$ & $\textbf{63\%}$ & $\textbf{1.118}$ & $\textbf{0.049}$ & $\textbf{80\%}$ \\
\hline
\end{tabular}
\label{table_MAP}
\end{table}

\begin{table}[t]
\centering
\caption{Quantitative comparison on the Famous dataset. The Famous dataset is only used for testing. The results show that our method achieves good generalizability and is robust to noise.}
\label{table2}
\begin{tabular}{ccccccc}
\hline
\multirow{2}{*}{Method} & \multicolumn{3}{c}{Famous no-noise} & \multicolumn{3}{c}{Famous var-noise}\cr
\cmidrule(lr){2-4} \cmidrule(lr){5-7}

  & CD & NCE & BCR & CD & NCE & BCR \\
\hline

CON  & $3.983$ & $0.241$ & 0\% & $4.189$ & $0.246$ &  0\%\\

GR  & $1.631$ & $0.103$ & 0\% & $1.706$ & $0.122$ &  0\%\\

SPR  & $1.359$ & $\textbf{0.081}$ & \textbf{68.2\%} & $1.380$ & $0.102$ & 22.7\% \\

P2S  & $1.235$  & $0.094$ & 0\% & $1.258$ & $0.102$ & 0\% \\

Ours  & $\textbf{1.226}$  & $0.082$ & 31.8\% & $\textbf{1.253}$ & $\textbf{0.091}$ & \textbf{77.3\%}\\
\hline
\end{tabular}
\label{table_MAP}
\end{table}

To demonstrate the efficacy of our approach, we first compare the results of our reconstructions with those of the state-of-the-art data-driven and non-data-driven methods mentioned in Section~\ref{experiments_overview} on the ABC test dataset and Famous dataset. For Points2Surf~\cite{2020Points2Surf}, we use the same training and test data as those in our network and apply uniform sub-samples, which is the best sampling method for them. We train at the same time in the same environment on the computing server with $4$ Tesla V$100$ SXM2 GPUs until both ours and theirs fully converge. All the following results are better than directly using the trained models they provided on our test data. For CON~\cite{2020ConvOccupancy}, when testing on the ABC test dataset, to achieve a fair comparison, we train their network on the ABC training dataset by using the default parameters. When experimenting on the Famous dataset, we directly use the model they trained on ShapeNet~\cite{2015ShapeNet}. For classical methods SPR~\cite{2013Screened} and GR~\cite{2019GaussRecon}, we utilize the recent PCPNet~\cite{2018PCPNet} with multi-scale architecture to estimate the normals for input. The no-noise version contains all clean data. For the var-noise version, we randomize the noise ratio and amplitude for each shape and add point-wise Gaussian noise as described earlier.

Quantitative comparisons in the ABC and Famous datasets are provided in Table~\ref{table1} and Table~\ref{table2}, respectively. Notably, we multiply the CD value by $100$. To demonstrate that normal consistency is a universal advantage of our method, i.e., our method is generally better, we calculate the \textbf{Best consistency rate} (BCR) of each method. BCR indicates the proportion of achieving the best normal consistency on a dataset. For example, the BCR of our method in the ABC var-noise dataset is $80\%$. That is, our method achieves the best normal consistency among all the methods in $80$ of the $100$ test shapes. The quantitative comparisons show that our method exhibits a comprehensive and well-rounded performance on both clean data and noisy data. Compared with data-driven methods CON and Points2Surf, our approach can generate surfaces with smaller CD and better normal consistency in all the test sets. Compared with traditional methods SPR and GR, our approach does not depend on normal estimation and is more robust to noise.

Fig.~\ref{fig:Figure6} and Fig.~\ref{fig:Figure7} illustrate the qualitative comparisons of different methods on the ABC no-noise and var-noise datasets respectively. Fig.~\ref{fig:Figure8} and Fig.~\ref{fig:Figure9} show the results on the Famous dataset. We can observe that CON fails to generate detailed geometric topology due to its plane projection with insufficient resolution. Many thin and regular objects in the ABC dataset can degenerate plane projection and produce poor results, making the average value on ABC clean data even worse than that on noisy data. GR requires dense sampling to reconstruct detailed information. It also relies heavily on normal accuracy and is considerably affected by inaccurate normal estimations, particularly of the orientation. SPR can generate high-quality results with accurate sampling and normals. However, it is also affected by inaccurate normal orientation and its performance degrades when sampling is noisy. Points2Surf mainly suffers from inconsistency and generates bumpy surfaces with ripples. Contrastingly, our method can generate high consistent surfaces without losing geometric details and sharp corners.

Notably, the normal consistency of SPR on the Famous no-noise dataset is outstanding. One reason for such result is that some of the training shapes used by PCPNet are selected from the Famous dataset. Therefore, the normal input of SPR is highly accurate for these shapes.

To verify the robustness of our method further, we also perform experiments at varying levels of noise amplitude on the Famous dataset. In these experiments, we add Gaussian noise to all the sample points. An amplitude $\beta$ indicates that we add point-wise Gaussian noise with a maximum value of $\beta$ and a standard deviation of $\frac{\beta}{3}$. We evaluate the CD and NCE of different methods on $\beta=0.01,0.02,0.04$. The results are presented in Table~\ref{table3}. Our method
achieves the best results in all three amplitudes. As noise increases, our method exhibits better robustness. To perceive the noise of the point cloud,  we should learn the local and overall shapes expressed by the point cloud. This information is compiled into local features and the global latent vector. We concatenate point features, local features and global latent vector in the network. Resultantly, we have enough information to perceive the noise scale of the point cloud. This explains why our method is more robust to noise.

\begin{table}[t]
\centering
\caption{Quantitative comparison on the Famous dataset with different noise amplitudes. n-0.01, n-0.02, and n-0.04 denotes the amplitude of Gaussian noise. Our method achieves the best results in all three amplitudes.}
\label{table3}
\begin{tabular}{ccccccc}
\hline
\multirow{2}{*}{Method} & \multicolumn{2}{c}{Famous n-0.01} & \multicolumn{2}{c}{Famous n-0.02}  & \multicolumn{2}{c}{Famous n-0.04} \cr
\cmidrule(lr){2-3} \cmidrule(lr){4-5} \cmidrule(lr){6-7}

  & CD & NCE & CD & NCE & CD & NCE \\
\hline

CON  & $3.989$ & $0.242$ & $4.137$ & $0.245$ & $4.570$ & $0.252$\\

GR  & $1.644$ & $0.110$ & $1.689$ & $0.125$ & $1.870$ &  $0.151$\\

SPR  & $1.365$ & $0.094$ & $1.382$ & $0.110$ & $1.441$ & $0.145$ \\

P2S  & $1.241$  & $0.097$ & $1.264$ & $0.105$ & $1.382$ & $0.143$ \\

Ours  & $\textbf{1.237}$  & $\textbf{0.086}$ & $\textbf{1.256}$ & $\textbf{0.094}$ & $\textbf{1.327}$ & $\textbf{0.117}$\\
\hline
\end{tabular}
\label{table_MAP}
\end{table}

To demonstrate that our method can adapt to complex geometric structures, we also conduct experiments on sculptures in the ThreeDScans dataset~\cite{3DScansWeb}. Fig.~\ref{fig:Figure10} visually compares the reconstruction results. Our method can generate available surfaces from raw point clouds even if the original shapes have complex geometric topology.

\begin{table}[t]
\centering
\caption{Ablation study of the neighbor size $k$ in the SEF Extractor.}
\label{table4}
\begin{tabular}{ccc}
\hline
\multirow{2}{*}{Method} & ABC var-noise & Famous var-noise\cr

  & NCE (relative) & NCE (relative) \\
\hline

No SEF Extractor & 1.12 & 1.05 \\

$k=5$  & 1.05 & 1.10 \\

$k=10$   & \textbf{1.00}  & \textbf{1.00} \\

$k=20$  & 1.18  & 1.09 \\
\hline
\end{tabular}
\label{table_MAP}
\end{table}
\subsection{Ablation study on SEF Extractor} \label{ablation_study}
To validate the effectiveness of the SEF Extractor, which we utilize to improve our learning of local properties, we conduct experiments on the neighbor size $k$ in the SEF Extractor.
Table~\ref{table4} provides the relative values of NCE with the same amount of training. If we eliminate the SEF Extractor, i.e., we only concatenate point features and global features, then we cannot obtain precise point-wise contributions and the accuracy is reduced. We perform experiments on $k=5$, $k=10$, and $k=20$, and determine that $k=10$ achieves the best results. The SEF Extractor is a symmetric sub-network, that is, the weights of the $k$ points are the same. Hence, an excessively large $k$ may be redundant and cause difficulties for network training and local details learning. Meanwhile, an extremely small $k$ is insufficient for distinguishing noise and negatively affects the local shape learning. Accordingly, we select $k=10$ in all the previous experiments.

\subsection{Comparison with explicit reconstructions}
\label{comparison_explict}

Most of explicit reconstruction methods use Delaunay triangles or Voronoi diagrams to reconstruct triangle meshes~\cite{2009Iso,2015Delaunay,2021LocalDelaunay}. They usually do not need normals either. However, explicit reconstruction methods usually tend to fit every sampling point. Thus, it is difficult for them to manage noisy inputs. In this section, we conduct a qualitative comparison with~\cite{2015Delaunay}, which defines a novel shape-hull graph and provides a Delaunay sculpting algorithm for closed surface reconstruction. We directly use the source code published by the authors with default parameters.

The results are shown in Fig.~\ref{fig:Figure11}. The first and the second examples are clean samplings and the third example is a noisy sampling. Delaunay sculpting can reconstruct high-quality meshes when the point clouds are clean and do not contain thin structures. However, its performance degenerates when thin structures exist or the input is noisy. Contrastingly, our method can handle these inputs. In the first example, we also generate the sharp edges better.

\subsection{Managing artifacts and sparse samplings} \label{Adaptability}
In addition to noise, other artifacts can also exist. For example, sampling point clouds may have holes caused by missing parts such as the bottom of a statue which cannot be scanned by the scanner. Thus, we want to close the holes caused by missing parts in the point clouds. For data-driven approaches, the best way to flexibly solve artifacts is to construct similar artifacts in the training data so that the network can learn to address this problem. Here, we take missing parts as a representative. We can also use the same approach to manage different types of artifacts. During training, we randomly add holes with a radius of [0,0.3] to 50 \% of the ABC training data and supervise them by the original indicator values. The model trained in this way can adapt well to point clouds with missing parts. We compare our methods with SPR~\cite{2013Screened} of two different normal estimation methods (PCA realized by Meshlab~\cite{2008Meshlab} and PCPNet proposed by Guerrero et al.~\cite{2018PCPNet}). The results are shown in Fig.~\ref{fig:Figure12}. Our method can effectively close the holes in the point clouds.

Besides, our requirements for sampling density may vary by task. In this section, we conduct experiments on sparsely sampled point clouds. We use the ABC dataset for training and Famous dataset for testing. The sampling points of the training data is [1000,5000].  In this experiment, we set the local patch size $n_d$ in the network to 30 and the neighbor size $k$ in the SEF Extractor to 5. Fig.~\ref{fig:Figure13} shows our reconstructions. We generate smooth surfaces with sparsely sampled point clouds. The sampling points of the two examples are 3131 and 2317. The point cloud in the ear of the second example is very sparse and the shape is relatively sharp, but our method can still tackle it well. This experiment shows that we can manage varying inputs as long as performing training set sampling according to our needs. When the sampling density of the test data significantly differs from the training set, we can not expect good results.

\section{Conclusion} \label{conclusion}
In this work, we introduce a deep learning approach for directly learning modified indicator functions from un-oriented point clouds. We apply the Gauss lemma to deep learning for the first time. Our network learns precise local properties, combines features of different scales, and performs surface integral by aggregating the point-wise contributions in the network. Sufficient experimental results demonstrate that our method can generate smooth and accurate reconstructions under noisy inputs. Moreover, it outperforms current state-of-the-art data-driven and non-data-driven approaches.

Our method still has some limitations, many of them are also common in deep learning approaches. Our method is not fast as it takes about 200 s to reconstruct a mesh in our environment. Nevertheless, the time complexity does not increase obviously with the point number and we do not need normals. Although our method can generate a relatively smoother surface compared with existing deep learning methods, predictions between query points are still independent of one another. This may result in some fragmented artifacts near the surface for some complex shapes. See the second example of Fig.~\ref{fig:Figure8}. For further improvement, we need to supervise the high-order properties of the implicit function, such as the gradient. Utilizing high-order properties between query points is also an interesting future direction for learning-based implicit representations.

\bibliographystyle{cag-num-names}
\bibliography{refs}

\begin{thebibliography}{38}
\providecommand{\natexlab}[1]{#1}
\providecommand{\url}[1]{\texttt{#1}}
\providecommand{\href}[2]{#2}
\providecommand{\path}[1]{#1}
\providecommand{\eprint}[1]{\href{http://arxiv.org/abs/#1}{\path{#1}}}
\providecommand{\DOIprefix}{doi:}
\providecommand{\ArXivprefix}{arXiv:}
\providecommand{\URLprefix}{URL: }
\providecommand{\Pubmedprefix}{pmid:}
\providecommand{\doi}[1]{\href{http://dx.doi.org/#1}{\path{#1}}}
\providecommand{\Pubmed}[1]{\href{pmid:#1}{\path{#1}}}
\providecommand{\BIBand}{and}
\providecommand{\bibinfo}[2]{#2}
\ifx\xfnm\undefined \def\xfnm[#1]{\unskip,\space#1}\fi
\bibitem[{Berger et~al.(2017)Berger, Tagliasacchi, Seversky, Alliez,
  Guennebaud, Levine et~al.}]{ReconSurvey2017}
\bibinfo{author}{Berger\xfnm[ M]}, \bibinfo{author}{Tagliasacchi\xfnm[ A]},
  \bibinfo{author}{Seversky\xfnm[ LM]}, \bibinfo{author}{Alliez\xfnm[ P]},
  \bibinfo{author}{Guennebaud\xfnm[ G]}, \bibinfo{author}{Levine\xfnm[ JA]},
  et~al.
\newblock \bibinfo{title}{A survey of surface reconstruction from point
  clouds}.
\newblock \bibinfo{journal}{Comput Graph Forum}
  \bibinfo{year}{2017};\bibinfo{volume}{36}(\bibinfo{number}{1}):\bibinfo{pages}{301--329}.
\bibitem[{Kazhdan(2005)}]{2005Recon}
\bibinfo{author}{Kazhdan\xfnm[ MM]}.
\newblock \bibinfo{title}{Reconstruction of solid models from oriented point
  sets}.
\newblock In: \bibinfo{editor}{Desbrun\xfnm[ M]},
  \bibinfo{editor}{Pottmann\xfnm[ H]}, editors. \bibinfo{booktitle}{Third
  Eurographics Symposium on Geometry Processing, Vienna, Austria, July 4-6,
  2005}; vol. \bibinfo{volume}{255} of \emph{\bibinfo{series}{{ACM}
  International Conference Proceeding Series}}.
  \bibinfo{publisher}{Eurographics Association}; \bibinfo{year}{2005}, p.
  \bibinfo{pages}{73--82}.
\bibitem[{Kazhdan et~al.(2006)Kazhdan, Bolitho and Hoppe}]{2006Poisson}
\bibinfo{author}{Kazhdan\xfnm[ MM]}, \bibinfo{author}{Bolitho\xfnm[ M]},
  \bibinfo{author}{Hoppe\xfnm[ H]}.
\newblock \bibinfo{title}{Poisson surface reconstruction}.
\newblock In: \bibinfo{editor}{Sheffer\xfnm[ A]},
  \bibinfo{editor}{Polthier\xfnm[ K]}, editors. \bibinfo{booktitle}{Proceedings
  of the Fourth Eurographics Symposium on Geometry Processing, Cagliari,
  Sardinia, Italy, June 26-28, 2006}; vol. \bibinfo{volume}{256}.
  \bibinfo{year}{2006}, p. \bibinfo{pages}{61--70}.
\bibitem[{Calakli and Taubin(2011)}]{2011SSD}
\bibinfo{author}{Calakli\xfnm[ F]}, \bibinfo{author}{Taubin\xfnm[ G]}.
\newblock \bibinfo{title}{{SSD:} smooth signed distance surface
  reconstruction}.
\newblock \bibinfo{journal}{Comput Graph Forum}
  \bibinfo{year}{2011};\bibinfo{volume}{30}(\bibinfo{number}{7}):\bibinfo{pages}{1993--2002}.
\bibitem[{Kazhdan and Hoppe(2013)}]{2013Screened}
\bibinfo{author}{Kazhdan\xfnm[ MM]}, \bibinfo{author}{Hoppe\xfnm[ H]}.
\newblock \bibinfo{title}{Screened poisson surface reconstruction}.
\newblock \bibinfo{journal}{{ACM} Trans Graph}
  \bibinfo{year}{2013};\bibinfo{volume}{32}(\bibinfo{number}{3}):\bibinfo{pages}{29:1--29:13}.
\bibitem[{Kazhdan et~al.(2020)Kazhdan, Chuang, Rusinkiewicz and
  Hoppe}]{2020EnvelopePoisson}
\bibinfo{author}{Kazhdan\xfnm[ M]}, \bibinfo{author}{Chuang\xfnm[ M]},
  \bibinfo{author}{Rusinkiewicz\xfnm[ S]}, \bibinfo{author}{Hoppe\xfnm[ H]}.
\newblock \bibinfo{title}{Poisson surface reconstruction with envelope
  constraints}.
\newblock \bibinfo{journal}{Comput Graph Forum}
  \bibinfo{year}{2020};\bibinfo{volume}{39}(\bibinfo{number}{5}):\bibinfo{pages}{173--182}.
\bibitem[{Lu et~al.(2019)Lu, Shi, Sun and Wang}]{2019GaussRecon}
\bibinfo{author}{Lu\xfnm[ W]}, \bibinfo{author}{Shi\xfnm[ Z]},
  \bibinfo{author}{Sun\xfnm[ J]}, \bibinfo{author}{Wang\xfnm[ B]}.
\newblock \bibinfo{title}{Surface reconstruction based on the modified gauss
  formula}.
\newblock \bibinfo{journal}{{ACM} Trans Graph}
  \bibinfo{year}{2019};\bibinfo{volume}{38}(\bibinfo{number}{1}):\bibinfo{pages}{2:1--2:18}.
\bibitem[{Erler et~al.(2020)Erler, Guerrero, Ohrhallinger, Mitra and
  Wimmer}]{2020Points2Surf}
\bibinfo{author}{Erler\xfnm[ P]}, \bibinfo{author}{Guerrero\xfnm[ P]},
  \bibinfo{author}{Ohrhallinger\xfnm[ S]}, \bibinfo{author}{Mitra\xfnm[ NJ]},
  \bibinfo{author}{Wimmer\xfnm[ M]}.
\newblock \bibinfo{title}{{Points2Surf}: Learning implicit surfaces from point
  clouds}.
\newblock In: \bibinfo{editor}{Vedaldi\xfnm[ A]},
  \bibinfo{editor}{Bischof\xfnm[ H]}, \bibinfo{editor}{Brox\xfnm[ T]},
  \bibinfo{editor}{Frahm\xfnm[ J]}, editors. \bibinfo{booktitle}{Computer
  Vision - {ECCV} 2020 - 16th European Conference, Glasgow, UK, August 23-28,
  2020, Proceedings, Part {V}}; vol. \bibinfo{volume}{12350}.
  \bibinfo{year}{2020}, p. \bibinfo{pages}{108--124}.
\bibitem[{Qi et~al.(2017{\natexlab{a}})Qi, Su, Mo and Guibas}]{2017PointNet}
\bibinfo{author}{Qi\xfnm[ CR]}, \bibinfo{author}{Su\xfnm[ H]},
  \bibinfo{author}{Mo\xfnm[ K]}, \bibinfo{author}{Guibas\xfnm[ LJ]}.
\newblock \bibinfo{title}{Pointnet: Deep learning on point sets for 3d
  classification and segmentation}.
\newblock In: \bibinfo{booktitle}{2017 {IEEE} Conference on Computer Vision and
  Pattern Recognition, {CVPR} 2017, Honolulu, HI, USA, July 21-26, 2017}.
  \bibinfo{year}{2017}{\natexlab{a}}, p. \bibinfo{pages}{77--85}.
\bibitem[{Qi et~al.(2017{\natexlab{b}})Qi, Yi, Su and Guibas}]{2017PointNet++}
\bibinfo{author}{Qi\xfnm[ CR]}, \bibinfo{author}{Yi\xfnm[ L]},
  \bibinfo{author}{Su\xfnm[ H]}, \bibinfo{author}{Guibas\xfnm[ LJ]}.
\newblock \bibinfo{title}{Pointnet++: Deep hierarchical feature learning on
  point sets in a metric space}.
\newblock In: \bibinfo{editor}{Guyon\xfnm[ I]}, \bibinfo{editor}{von
  Luxburg\xfnm[ U]}, \bibinfo{editor}{Bengio\xfnm[ S]},
  \bibinfo{editor}{Wallach\xfnm[ HM]}, \bibinfo{editor}{Fergus\xfnm[ R]},
  \bibinfo{editor}{Vishwanathan\xfnm[ SVN]}, et~al., editors.
  \bibinfo{booktitle}{Advances in Neural Information Processing Systems 30:
  Annual Conference on Neural Information Processing Systems 2017, December
  4-9, 2017, Long Beach, CA, {USA}}. \bibinfo{year}{2017}{\natexlab{b}}, p.
  \bibinfo{pages}{5099--5108}.
\bibitem[{Li et~al.(2018)Li, Bu, Sun, Wu, Di and Chen}]{2018PointCNN}
\bibinfo{author}{Li\xfnm[ Y]}, \bibinfo{author}{Bu\xfnm[ R]},
  \bibinfo{author}{Sun\xfnm[ M]}, \bibinfo{author}{Wu\xfnm[ W]},
  \bibinfo{author}{Di\xfnm[ X]}, \bibinfo{author}{Chen\xfnm[ B]}.
\newblock \bibinfo{title}{Pointcnn: Convolution on x-transformed points}.
\newblock In: \bibinfo{editor}{Bengio\xfnm[ S]}, \bibinfo{editor}{Wallach\xfnm[
  HM]}, \bibinfo{editor}{Larochelle\xfnm[ H]}, \bibinfo{editor}{Grauman\xfnm[
  K]}, \bibinfo{editor}{Cesa{-}Bianchi\xfnm[ N]},
  \bibinfo{editor}{Garnett\xfnm[ R]}, editors. \bibinfo{booktitle}{Advances in
  Neural Information Processing Systems 31: Annual Conference on Neural
  Information Processing Systems 2018, NeurIPS 2018, December 3-8, 2018,
  Montr{\'{e}}al, Canada}. \bibinfo{year}{2018}, p. \bibinfo{pages}{828--838}.
\bibitem[{Wu et~al.(2019)Wu, Qi and Li}]{2019PointConv}
\bibinfo{author}{Wu\xfnm[ W]}, \bibinfo{author}{Qi\xfnm[ Z]},
  \bibinfo{author}{Li\xfnm[ F]}.
\newblock \bibinfo{title}{Pointconv: Deep convolutional networks on 3d point
  clouds}.
\newblock In: \bibinfo{booktitle}{{IEEE} Conference on Computer Vision and
  Pattern Recognition, {CVPR} 2019, Long Beach, CA, USA, June 16-20, 2019}.
  \bibinfo{year}{2019}, p. \bibinfo{pages}{9621--9630}.
\bibitem[{Guerrero et~al.(2018)Guerrero, Kleiman, Ovsjanikov and
  Mitra}]{2018PCPNet}
\bibinfo{author}{Guerrero\xfnm[ P]}, \bibinfo{author}{Kleiman\xfnm[ Y]},
  \bibinfo{author}{Ovsjanikov\xfnm[ M]}, \bibinfo{author}{Mitra\xfnm[ NJ]}.
\newblock \bibinfo{title}{{PCPNet}: Learning local shape properties from raw
  point clouds}.
\newblock \bibinfo{journal}{Comput Graph Forum}
  \bibinfo{year}{2018};\bibinfo{volume}{37}(\bibinfo{number}{2}):\bibinfo{pages}{75--85}.
\bibitem[{Groueix et~al.(2018)Groueix, Fisher, Kim, Russell and
  Aubry}]{2018AtlasNet}
\bibinfo{author}{Groueix\xfnm[ T]}, \bibinfo{author}{Fisher\xfnm[ M]},
  \bibinfo{author}{Kim\xfnm[ VG]}, \bibinfo{author}{Russell\xfnm[ B]},
  \bibinfo{author}{Aubry\xfnm[ M]}.
\newblock \bibinfo{title}{{AtlasNet: A Papier-M\^ach\'e Approach to Learning 3D
  Surface Generation}}.
\newblock In: \bibinfo{booktitle}{Proceedings IEEE Conf. on Computer Vision and
  Pattern Recognition (CVPR)}. \bibinfo{year}{2018},.
\bibitem[{Mescheder et~al.(2019)Mescheder, Oechsle, Niemeyer, Nowozin and
  Geiger}]{2019Occupancy}
\bibinfo{author}{Mescheder\xfnm[ LM]}, \bibinfo{author}{Oechsle\xfnm[ M]},
  \bibinfo{author}{Niemeyer\xfnm[ M]}, \bibinfo{author}{Nowozin\xfnm[ S]},
  \bibinfo{author}{Geiger\xfnm[ A]}.
\newblock \bibinfo{title}{Occupancy networks: Learning 3d reconstruction in
  function space}.
\newblock In: \bibinfo{booktitle}{{IEEE} Conference on Computer Vision and
  Pattern Recognition, {CVPR} 2019, Long Beach, CA, USA, June 16-20, 2019}.
  \bibinfo{year}{2019}, p. \bibinfo{pages}{4460--4470}.
\bibitem[{Peng et~al.(2020)Peng, Niemeyer, Mescheder, Pollefeys and
  Geiger}]{2020ConvOccupancy}
\bibinfo{author}{Peng\xfnm[ S]}, \bibinfo{author}{Niemeyer\xfnm[ M]},
  \bibinfo{author}{Mescheder\xfnm[ LM]}, \bibinfo{author}{Pollefeys\xfnm[ M]},
  \bibinfo{author}{Geiger\xfnm[ A]}.
\newblock \bibinfo{title}{Convolutional occupancy networks}.
\newblock In: \bibinfo{editor}{Vedaldi\xfnm[ A]},
  \bibinfo{editor}{Bischof\xfnm[ H]}, \bibinfo{editor}{Brox\xfnm[ T]},
  \bibinfo{editor}{Frahm\xfnm[ J]}, editors. \bibinfo{booktitle}{Computer
  Vision - {ECCV} 2020 - 16th European Conference, Glasgow, UK, August 23-28,
  2020, Proceedings, Part {III}}; vol. \bibinfo{volume}{12348} of
  \emph{\bibinfo{series}{Lecture Notes in Computer Science}}.
  \bibinfo{year}{2020}, p. \bibinfo{pages}{523--540}.
\bibitem[{Park et~al.(2019)Park, Florence, Straub, Newcombe and
  Lovegrove}]{2019DeepSDF}
\bibinfo{author}{Park\xfnm[ JJ]}, \bibinfo{author}{Florence\xfnm[ P]},
  \bibinfo{author}{Straub\xfnm[ J]}, \bibinfo{author}{Newcombe\xfnm[ RA]},
  \bibinfo{author}{Lovegrove\xfnm[ S]}.
\newblock \bibinfo{title}{Deepsdf: Learning continuous signed distance
  functions for shape representation}.
\newblock In: \bibinfo{booktitle}{{IEEE} Conference on Computer Vision and
  Pattern Recognition, {CVPR} 2019, Long Beach, CA, USA, June 16-20, 2019}.
  \bibinfo{year}{2019}, p. \bibinfo{pages}{165--174}.
\bibitem[{Dai and Nie{\ss}ner(2019)}]{2019Scan2Mesh}
\bibinfo{author}{Dai\xfnm[ A]}, \bibinfo{author}{Nie{\ss}ner\xfnm[ M]}.
\newblock \bibinfo{title}{Scan2mesh: From unstructured range scans to 3d
  meshes}.
\newblock In: \bibinfo{booktitle}{{IEEE} Conference on Computer Vision and
  Pattern Recognition, {CVPR} 2019, Long Beach, CA, USA, June 16-20, 2019}.
  \bibinfo{year}{2019}, p. \bibinfo{pages}{5574--5583}.
\bibitem[{Folland(1995)}]{1995PDE}
\bibinfo{author}{Folland\xfnm[ GB]}.
\newblock \bibinfo{title}{Introduction to partial differential equations}.
\newblock \bibinfo{journal}{Texts in Applied Mathematics}
  \bibinfo{year}{1995};.
\bibitem[{Wendland(2009)}]{2009Potential}
\bibinfo{author}{Wendland\xfnm[ WL]}.
\newblock \bibinfo{title}{On the Double Layer Potential}.
\newblock \bibinfo{publisher}{Birkhäuser Basel}; \bibinfo{year}{2009}.
\bibitem[{Hoppe et~al.(1992)Hoppe, DeRose, Duchamp, McDonald and
  Stuetzle}]{1992UnorganizedRecon}
\bibinfo{author}{Hoppe\xfnm[ H]}, \bibinfo{author}{DeRose\xfnm[ T]},
  \bibinfo{author}{Duchamp\xfnm[ T]}, \bibinfo{author}{McDonald\xfnm[ JA]},
  \bibinfo{author}{Stuetzle\xfnm[ W]}.
\newblock \bibinfo{title}{Surface reconstruction from unorganized points}.
\newblock In: \bibinfo{editor}{Thomas\xfnm[ JJ]}, editor.
  \bibinfo{booktitle}{Proceedings of the 19th Annual Conference on Computer
  Graphics and Interactive Techniques, {SIGGRAPH} 1992, Chicago, IL, USA, July
  27-31}. \bibinfo{year}{1992}, p. \bibinfo{pages}{71--78}.
\bibitem[{Fuhrmann and Goesele(2014)}]{2014FloatingScale}
\bibinfo{author}{Fuhrmann\xfnm[ S]}, \bibinfo{author}{Goesele\xfnm[ M]}.
\newblock \bibinfo{title}{Floating scale surface reconstruction}.
\newblock \bibinfo{journal}{{ACM} Trans Graph}
  \bibinfo{year}{2014};\bibinfo{volume}{33}(\bibinfo{number}{4}):\bibinfo{pages}{46:1--46:11}.
\bibitem[{Huang et~al.(2019)Huang, Carr and Ju}]{2019VIPSS}
\bibinfo{author}{Huang\xfnm[ Z]}, \bibinfo{author}{Carr\xfnm[ N]},
  \bibinfo{author}{Ju\xfnm[ T]}.
\newblock \bibinfo{title}{Variational implicit point set surfaces}.
\newblock \bibinfo{journal}{{ACM} Trans Graph}
  \bibinfo{year}{2019};\bibinfo{volume}{38}(\bibinfo{number}{4}):\bibinfo{pages}{124:1--124:13}.
\bibitem[{Chabra et~al.(2020)Chabra, Lenssen, Ilg, Schmidt, Straub, Lovegrove
  et~al.}]{2020DeepLocalShapes}
\bibinfo{author}{Chabra\xfnm[ R]}, \bibinfo{author}{Lenssen\xfnm[ JE]},
  \bibinfo{author}{Ilg\xfnm[ E]}, \bibinfo{author}{Schmidt\xfnm[ T]},
  \bibinfo{author}{Straub\xfnm[ J]}, \bibinfo{author}{Lovegrove\xfnm[ S]},
  et~al.
\newblock \bibinfo{title}{Deep local shapes: Learning local {SDF} priors for
  detailed 3d reconstruction}.
\newblock In: \bibinfo{booktitle}{Computer Vision - {ECCV} 2020 - 16th European
  Conference, Glasgow, UK, August 23-28, 2020, Proceedings, Part {XXIX}}; vol.
  \bibinfo{volume}{12374}. \bibinfo{year}{2020}, p. \bibinfo{pages}{608--625}.
\bibitem[{Atzmon and Lipman(2020)}]{SAL2020}
\bibinfo{author}{Atzmon\xfnm[ M]}, \bibinfo{author}{Lipman\xfnm[ Y]}.
\newblock \bibinfo{title}{{SAL:} sign agnostic learning of shapes from raw
  data}.
\newblock In: \bibinfo{booktitle}{2020 {IEEE/CVF} Conference on Computer Vision
  and Pattern Recognition, {CVPR} 2020, Seattle, WA, USA, June 13-19, 2020}.
  \bibinfo{publisher}{{IEEE}}; \bibinfo{year}{2020}, p.
  \bibinfo{pages}{2562--2571}.
\bibitem[{Zhao et~al.(2020)Zhao, Lei, Wen, Zhang and Jia}]{SALRecon2020}
\bibinfo{author}{Zhao\xfnm[ W]}, \bibinfo{author}{Lei\xfnm[ J]},
  \bibinfo{author}{Wen\xfnm[ Y]}, \bibinfo{author}{Zhang\xfnm[ J]},
  \bibinfo{author}{Jia\xfnm[ K]}.
\newblock \bibinfo{title}{Sign-agnostic implicit learning of surface
  self-similarities for shape modeling and reconstruction from raw point
  clouds}.
\newblock \bibinfo{journal}{CoRR}
  \bibinfo{year}{2020};\bibinfo{volume}{abs/2012.07498}.
\newblock \URLprefix \url{https://arxiv.org/abs/2012.07498}.
  \href{http://arxiv.org/abs/2012.07498}{\tt arXiv:2012.07498}.
\bibitem[{Jaderberg et~al.(2015)Jaderberg, Simonyan, Zisserman and
  Kavukcuoglu}]{STN15}
\bibinfo{author}{Jaderberg\xfnm[ M]}, \bibinfo{author}{Simonyan\xfnm[ K]},
  \bibinfo{author}{Zisserman\xfnm[ A]}, \bibinfo{author}{Kavukcuoglu\xfnm[ K]}.
\newblock \bibinfo{title}{Spatial transformer networks}.
\newblock In: \bibinfo{editor}{Cortes\xfnm[ C]},
  \bibinfo{editor}{Lawrence\xfnm[ ND]}, \bibinfo{editor}{Lee\xfnm[ DD]},
  \bibinfo{editor}{Sugiyama\xfnm[ M]}, \bibinfo{editor}{Garnett\xfnm[ R]},
  editors. \bibinfo{booktitle}{Advances in Neural Information Processing
  Systems 28: Annual Conference on Neural Information Processing Systems 2015,
  December 7-12, 2015, Montreal, Quebec, Canada}. \bibinfo{year}{2015}, p.
  \bibinfo{pages}{2017--2025}.
\bibitem[{Peethambaran and Muthuganapathy(2015)}]{2015Delaunay}
\bibinfo{author}{Peethambaran\xfnm[ J]}, \bibinfo{author}{Muthuganapathy\xfnm[
  R]}.
\newblock \bibinfo{title}{Reconstruction of water-tight surfaces through
  delaunay sculpting}.
\newblock \bibinfo{journal}{Comput Aided Des}
  \bibinfo{year}{2015};\bibinfo{volume}{58}:\bibinfo{pages}{62--72}.
\bibitem[{Huang et~al.(2009)Huang, Li, Zhang, Ascher and
  Cohen{-}Or}]{2009Consolidation}
\bibinfo{author}{Huang\xfnm[ H]}, \bibinfo{author}{Li\xfnm[ D]},
  \bibinfo{author}{Zhang\xfnm[ H]}, \bibinfo{author}{Ascher\xfnm[ UM]},
  \bibinfo{author}{Cohen{-}Or\xfnm[ D]}.
\newblock \bibinfo{title}{Consolidation of unorganized point clouds for surface
  reconstruction}.
\newblock \bibinfo{journal}{{ACM} Trans Graph}
  \bibinfo{year}{2009};\bibinfo{volume}{28}(\bibinfo{number}{5}):\bibinfo{pages}{176}.
\bibitem[{Lorensen and Cline(1987)}]{1987MarchingCubes}
\bibinfo{author}{Lorensen\xfnm[ WE]}, \bibinfo{author}{Cline\xfnm[ HE]}.
\newblock \bibinfo{title}{Marching cubes: {A} high resolution 3d surface
  construction algorithm}.
\newblock In: \bibinfo{editor}{Stone\xfnm[ MC]}, editor.
  \bibinfo{booktitle}{Proceedings of the 14th Annual Conference on Computer
  Graphics and Interactive Techniques, {SIGGRAPH} 1987, Anaheim, California,
  USA, July 27-31, 1987}. \bibinfo{year}{1987}, p. \bibinfo{pages}{163--169}.
\bibitem[{Lewiner et~al.(2003)Lewiner, Lopes, Vieira and
  Tavares}]{2013EfficientMarching}
\bibinfo{author}{Lewiner\xfnm[ T]}, \bibinfo{author}{Lopes\xfnm[ H]},
  \bibinfo{author}{Vieira\xfnm[ AW]}, \bibinfo{author}{Tavares\xfnm[ G]}.
\newblock \bibinfo{title}{Efficient implementation of marching cubes' cases
  with topological guarantees}.
\newblock \bibinfo{journal}{J Graphics, GPU, {\&} Game Tools}
  \bibinfo{year}{2003};\bibinfo{volume}{8}(\bibinfo{number}{2}):\bibinfo{pages}{1--15}.
\bibitem[{Koch et~al.(2019)Koch, Matveev, Jiang, Williams, Artemov, Burnaev
  et~al.}]{2019ABC}
\bibinfo{author}{Koch\xfnm[ S]}, \bibinfo{author}{Matveev\xfnm[ A]},
  \bibinfo{author}{Jiang\xfnm[ Z]}, \bibinfo{author}{Williams\xfnm[ F]},
  \bibinfo{author}{Artemov\xfnm[ A]}, \bibinfo{author}{Burnaev\xfnm[ E]},
  et~al.
\newblock \bibinfo{title}{{ABC:} {A} big {CAD} model dataset for geometric deep
  learning}.
\newblock In: \bibinfo{booktitle}{{IEEE} Conference on Computer Vision and
  Pattern Recognition, {CVPR} 2019, Long Beach, CA, USA, June 16-20, 2019}.
  \bibinfo{year}{2019}, p. \bibinfo{pages}{9601--9611}.
\bibitem[{Albertina et~al.(2021)Albertina, Museum, Museum, Guimet, Paris
  Musée~des Monuments~français, des sculptures de~la Ville~de
  et~al.}]{3DScansWeb}
\bibinfo{author}{Albertina\xfnm[ V]}, \bibinfo{author}{Museum\xfnm[ VK]},
  \bibinfo{author}{Museum\xfnm[ VT]}, \bibinfo{author}{Guimet\xfnm[ PM]},
  \bibinfo{author}{Paris Musée~des Monuments~français\xfnm[ Cdledp]},
  \bibinfo{author}{des sculptures de~la Ville~de\xfnm[ PD]}, et~al.
\newblock \bibinfo{title}{Three d scans}.
\newblock \bibinfo{year}{Accessed: 2021},\URLprefix
  \url{https://threedscans.com}.
\bibitem[{{Dawson Haggerty et al.}(2021)}]{2019trimesh}
\bibinfo{author}{{Dawson Haggerty et al.}\xfnm[]}.
\newblock \bibinfo{title}{trimesh 3.9}.
\newblock \bibinfo{year}{2021}.
\newblock \URLprefix \url{https://trimsh.org/}.
\bibitem[{Chang et~al.(2015)Chang, Funkhouser, Guibas, Hanrahan, Huang, Li
  et~al.}]{2015ShapeNet}
\bibinfo{author}{Chang\xfnm[ AX]}, \bibinfo{author}{Funkhouser\xfnm[ TA]},
  \bibinfo{author}{Guibas\xfnm[ LJ]}, \bibinfo{author}{Hanrahan\xfnm[ P]},
  \bibinfo{author}{Huang\xfnm[ Q]}, \bibinfo{author}{Li\xfnm[ Z]}, et~al.
\newblock \bibinfo{title}{Shapenet: An information-rich 3d model repository}.
\newblock \bibinfo{journal}{CoRR}
  \bibinfo{year}{2015};\bibinfo{volume}{abs/1512.03012}.
\newblock \URLprefix \url{http://arxiv.org/abs/1512.03012}.
  \href{http://arxiv.org/abs/1512.03012}{\tt arXiv:1512.03012}.
\bibitem[{Dey et~al.(2009)Dey, Li, Ramos and Wenger}]{2009Iso}
\bibinfo{author}{Dey\xfnm[ TK]}, \bibinfo{author}{Li\xfnm[ K]},
  \bibinfo{author}{Ramos\xfnm[ EA]}, \bibinfo{author}{Wenger\xfnm[ R]}.
\newblock \bibinfo{title}{Isotopic reconstruction of surfaces with boundaries}.
\newblock \bibinfo{journal}{Comput Graph Forum}
  \bibinfo{year}{2009};\bibinfo{volume}{28}(\bibinfo{number}{5}):\bibinfo{pages}{1371--1382}.
\bibitem[{Thayyil et~al.(2021)Thayyil, Yadav, Polthier and
  Muthuganapathy}]{2021LocalDelaunay}
\bibinfo{author}{Thayyil\xfnm[ SB]}, \bibinfo{author}{Yadav\xfnm[ SK]},
  \bibinfo{author}{Polthier\xfnm[ K]}, \bibinfo{author}{Muthuganapathy\xfnm[
  R]}.
\newblock \bibinfo{title}{Local delaunay-based high fidelity surface
  reconstruction from 3d point sets}.
\newblock \bibinfo{journal}{Comput Aided Geom Des}
  \bibinfo{year}{2021};\bibinfo{volume}{86}:\bibinfo{pages}{101973}.
\bibitem[{Cignoni et~al.(2008)Cignoni, Callieri, Corsini, Dellepiane, Ganovelli
  and Ranzuglia}]{2008Meshlab}
\bibinfo{author}{Cignoni\xfnm[ P]}, \bibinfo{author}{Callieri\xfnm[ M]},
  \bibinfo{author}{Corsini\xfnm[ M]}, \bibinfo{author}{Dellepiane\xfnm[ M]},
  \bibinfo{author}{Ganovelli\xfnm[ F]}, \bibinfo{author}{Ranzuglia\xfnm[ G]}.
\newblock \bibinfo{title}{Meshlab: an open-source mesh processing tool}.
\newblock In: \bibinfo{editor}{Scarano\xfnm[ V]}, \bibinfo{editor}{Chiara\xfnm[
  RD]}, \bibinfo{editor}{Erra\xfnm[ U]}, editors.
  \bibinfo{booktitle}{Eurographics Italian Chapter Conference 2008, Salerno,
  Italy, 2008}. \bibinfo{publisher}{Eurographics}; \bibinfo{year}{2008}, p.
  \bibinfo{pages}{129--136}.

\end{thebibliography}

\end{document}